%% file: camera_ready.tex
\documentclass[journal]{IEEEtran}

\input{config}

\title{Integration of Robot and Scene Kinematics for Sequential Mobile Manipulation Planning}

\author{Ziyuan~Jiao,~\IEEEmembership{Member,~IEEE}, 
        Yida~Niu,~\IEEEmembership{Student Member,~IEEE}, 
        Zeyu~Zhang,~\IEEEmembership{Member,~IEEE}, 
        Yangyang~Wu,
        Yao~Su,~\IEEEmembership{Member,~IEEE},
        Yixin~Zhu,~\IEEEmembership{Member,~IEEE}, 
        Hangxin~Liu,~\IEEEmembership{Member,~IEEE}
        and Song-Chun~Zhu,~\IEEEmembership{Fellow,~IEEE}
        
\thanks{\textit{Corresponding author: Hangxin Liu.}} 
   
\thanks{This article is an extended version of IEEE/RSJ International Conference on Intelligent Robots and Systems (IROS), Oct., 2021~\cite{jiao2021consolidated,jiao2021efficient}.}

\thanks{This work was supported in part by the National Natural Science Foundation of China (Grant No.52305007).}

\thanks{Ziyuao Jiao, Yida Niu, Zeyu Zhang, Yangyang Wu, Yao Su, Hangxin Liu, and Song-Chun Zhu are with State Key Laboratory of General Artificial Intelligence, Beijing Institute for General Artificial Intelligence (BIGAI), Beijing 100080, China (emails: jiaoziyuan@bigai.ai; niuyida@bigai.ai; zhangzeyu@bigai.ai; wuyangyang@bigai.ai; suyao@bigai.ai; liuhx@bigai.ai; sczhu@bigai.ai).} 

\thanks{Yida Niu is also with Institute for Artificial Intelligence, Peking University, Beijing 100871, China.}

\thanks{Yixin Zhu is with the School of Psychological and Cognitive Sciences, and the Institute for Artificial Intelligence, Peking University, Beijing 100871, China (email: yixin.zhu@pku.edu.cn).}

\thanks{Song-Chun Zhu is also with Institute for Artificial Intelligence and School of Artificial Intelligence, Peking University, Beijing 100871, China, and also with Department of Automation, Tsinghua University, Beijing 100084, China.}}

\begin{document}

\maketitle


\begin{abstract}
    We present a Sequential Mobile Manipulation Planning (SMMP) framework that can solve long-horizon multi-step mobile manipulation tasks with coordinated whole-body motion, even when interacting with articulated objects. By abstracting environmental structures as kinematic models and integrating them with the robot's kinematics, we construct an Augmented Configuration Apace (A-Space) that unifies the previously separate task constraints for navigation and manipulation, while accounting for the joint reachability of the robot base, arm, and manipulated objects. This integration facilitates efficient planning within a tri-level framework: a task planner generates symbolic action sequences to model the evolution of A-Space, an optimization-based motion planner computes continuous trajectories within A-Space to achieve desired configurations for both the robot and scene elements, and an intermediate plan refinement stage selects action goals that ensure long-horizon feasibility. Our simulation studies first confirm that planning in A-Space achieves an 84.6\% higher task success rate compared to baseline methods. Validation on real robotic systems demonstrates fluid mobile manipulation involving (i) seven types of rigid and articulated objects across 17 distinct contexts, and (ii) long-horizon tasks of up to 14 sequential steps. Our results highlight the significance of modeling scene kinematics into planning entities, rather than encoding task-specific constraints, offering a scalable and generalizable approach to complex robotic manipulation.
\end{abstract}
    
\begin{IEEEkeywords}
    Sequential mobile manipulation planning, kinematics, trajectory optimization, and service robot. \vspace{-6pt}
\end{IEEEkeywords}

\section{Introduction}\label{sec:introduction}

\IEEEPARstart{A}{utonomous} robots are increasingly being integrated into diverse environments in human society. Whether assisting people in daily activities~\cite{billard2019trends,kroemer2021review} or operating in outposts such as space stations or extraterrestrial bases~\cite{diffler2003human,jiang2022progress}, robot operations face significant challenges in performing sequential mobile manipulation tasks that require a range of manipulation skills and the ability to sequence these skills in expansive workspaces.

\begin{figure*}[t!]
    \centering
    \includegraphics[width=0.95\linewidth]{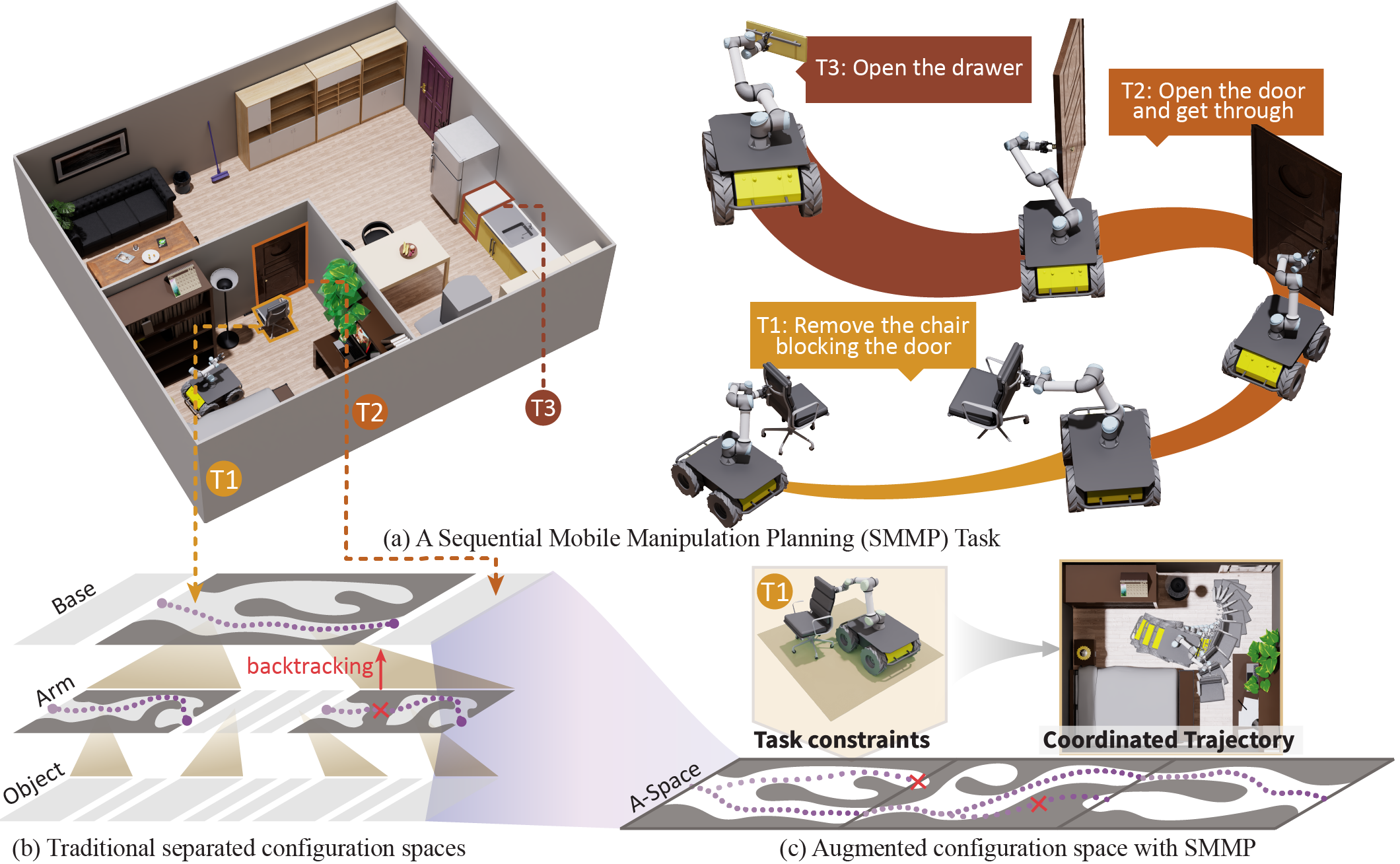}
    \caption{\textbf{An exemplar household task illustrating the advancement of proposed \ac{smmp} framework with A-Space compared to traditional planning.} (a) In this long-horizon task, the robot must (T1) remove the chair to approach the bedroom door, (T2) open the door and pass through it, and (T3) open the kitchen drawer. (b) Separated base-arm-object planning faces inherent challenges: as the base moves, the arm and object's configuration space evolves, frequently making initially feasible trajectories infeasible. This approach simplifies the planning complexity but may require costly backtracking across different configuration spaces, particularly in tasks demanding whole-body coordination. (c) By leveraging A-Space, separate configuration spaces are unified through \ac{akr}, incorporating various task constraints to enable coordinated base-arm-object trajectories across multiple steps via trajectory optimization, reducing the need for hierarchical backtracking.}
    \label{fig:overview}
    \vspace{-12pt}
\end{figure*}

\cref{fig:overview}(a) illustrates a typical \acf{smmp} scenario. Operating in cluttered workspaces poses significant challenges due to complex obstacle configurations~\cite{marcucci2023motion}. Robots are often required to balance both navigation and manipulation, \ie, mobile manipulation, to accomplish their goals~\cite{khatib1999robots,brock2016mobility}. Moreover, contact with diverse structures and objects introduces a wide range of task objectives and constraints, which are difficult to emulate in advance, particularly when dealing with articulated objects~\cite{karayiannidis2016adaptive,martin2019coupled,jiao2021consolidated}. Compounding this difficulty, robot actions can change the environment in ways that hinder the feasibility of future steps in long-horizon tasks~\cite{kaelbling2011hierarchical,toussaint2015logic,jiao2022sequential,sleiman2023versatile}. Therefore, the successful execution of an action in long-horizon mobile manipulation requires not only coordinating base-arm-object trajectories for individual steps, but also reasoning about the long-term implications of each action on future task feasibility.

Achieving coordinated trajectories of the robot’s base, arm, and manipulated object can become computationally intractable in long-horizon tasks due to the inherently interdependent configuration spaces during interactions, as shown in \cref{fig:overview}(b). Consequently, the likelihood of finding connected feasible paths across consecutive steps is low, and costly backtracking is often required when the planner encounters dead ends. A hierarchical strategy is typically employed to decompose the task execution into a sequence of primitive motions~\cite{kaelbling2011hierarchical,lozano2014constraint,jiao2022sequential}, facilitating more efficient trajectory generation and reducing computation costs in the face of errors. 
However, current hierarchical methods such as \ac{tamp} are primarily effective only for pick-and-place tasks~\cite{dantam2018task,garrett2020integrated,kim2022representation}, failing to scale to complex mobile manipulation tasks.
This limitation arises because complex mobile manipulation tasks require tightly coordinated navigation and manipulation, which are difficult to express symbolically. Semantic symbols often fail to capture critical geometric constraints necessary for task success, such as valid base positioning, interdependent base and arm movements, and collision avoidance. For example, the tasks in \cref{fig:overview} involve coupled base-arm-object interactions that would demand an intractable number of symbolic predicates to model accurately. 

In stark contrast, humans exhibit fluid manipulation skills and interact adeptly with their environment. Theories in cognitive psychology and philosophy suggest the concept of body schema: humans maintain a flexible representation of their bodies, enabling them to treat manipulated objects as extensions of their limbs during interactions~\cite{gallagher2006body,holmes2006beyond}. Embodied cognition studies further highlight that human intelligence is deeply intertwined with the environment~\cite{clark1999towards,wilson2002six}. 

Drawing from these insights, we propose to treat the environment and the robot embodiment as a whole for an efficient solution to mobile manipulation tasks. Specifically, we consolidate kinematic abstractions of scene elements~\cite{han2022scene}, robot arm, and navigational movements into an \acf{akr}. This consolidation merges originally separated yet entangled robot and object's configuration spaces into a single one, which we termed Augmented Configuration Space (A-Space), as depicted in \cref{fig:overview}(c). From this perspective, planning sequential mobile manipulation in A-Space can jointly account for the reachability of the robot base, arm, and manipulated objects and their inherent motion constraints, thus better ensuring spatial feasibility of sophisticated robot movements and resolving action temporal dependency among long-horizon tasks.

As A-Space is highly complex due to the extended \acp{dof}, we design a tri-level planning framework for efficient planning within it. By formulating an optimization-based motion planner, we can compute continuous trajectories to reach the desired configurations of both robot and scene entities, resulting in coordinated whole-body motions that satisfy relevant constraints. Furthermore, we design a task planner that defines symbolic states more intuitively and models actions along with their effects on A-Space, thereby enabling validation of motion feasibility over extended horizons by traversing temporally successive configuration spaces. Extending our previous foundation in task planning and motion planning~\cite{jiao2021consolidated,jiao2021efficient}, a newly designed plan refinement algorithm further combines them to resolve potential conflicts between anticipated future actions. Together, these components form a scalable \ac{smmp} framework capable of handling long-horizon mobile manipulation tasks involving diverse interactions within complex environments. We demonstrate the framework's effectiveness on a real robot system by performing coordinated base-arm-object motions across 17 diverse scenarios involving complex environmental structures. Moreover, the system successfully completes a 14-step long-horizon mobile manipulation task in a cluttered living room, with each step characterized by unique contact configurations. Additional simulation studies further validate our approach, quantifying improvements in execution efficiency and planning success rates when using A-Space compared to traditional planning paradigms.

Our contribution is fourfold:
\begin{enumerate}[leftmargin=*]
    \item We introduce an \ac{smmp} framework that solves long-horizon mobile manipulation tasks, with a newly proposed plan refinement algorithm that considers future actions while generating the motion planning problem for the current action, effectively increasing task success rates in complex long-horizon \ac{smmp} problems.
    \item We model the mobile manipulation planning problem from the \ac{akr} perspective, formulating the mobile manipulation planning problem as a trajectory optimization problem within the A-Space that integrates task specifications.
    \item We design a \ac{pddl}-based task planning domain describing the evolution of the A-Space, generalizing it to various daily long-horizon indoor mobile manipulation tasks.
    \item Through simulations, we validate the proposed method, achieving an 84.6\% improvement in success rate over baseline methods. With extensive experiments on physical mobile manipulators, we demonstrate the proposed method's feasibility across 7 types of rigid and articulated objects in 17 different contexts, with long-horizon tasks involving up to 14 steps.
\end{enumerate}

\subsection{Overview}
The remainder of this article is organized as follows. 
\cref{sec:review} reviews the literature and compares existing research with the contributions of this work. \cref{sec:modeling} introduces the proposed \ac{akr}-based modeling method for mobile manipulation. 
Based on the idea of \ac{akr}, \cref{sec:planning} formulates the corresponding motion planning and task planning setups, and \cref{sec:refine} elaborates the newly proposed plan refinement algorithm that bridges \ac{akr}-based motion planning with task planning components into a coherent \ac{smmp} system. 
Finally, \cref{sec:sim} and \cref{sec:exp} demonstrate the efficacy of \acp{akr} through simulations and experiments, respectively. \cref{sec:conclusion} concludes the paper with an in-depth discussion of key findings and future directions.

\section{Related Work}\label{sec:review}
\subsection{Mobile Manipulation}
Recently, notable efforts have focused on algorithms and system implementations to coordinate navigation and manipulation for mobile manipulation, especially within household environments. For instance, graph search~\cite{chitta2010planning}, equilibrium point control~\cite{jain2010pulling}, adaptive control~\cite{karayiannidis2016adaptive}, impedance control~\cite{stuede2019door}, and model predictive control~\cite{minniti2019whole} have been introduced for tasks like opening doors and drawers. For object retrieval or relocation in confined and cluttered spaces, methods such as the coevolutionary algorithm in~\cite{berenson2008optimization}, which jointly optimizes grasping and base poses, and adaptive dimensionality reduction in~\cite{gochev2012planning}, which manages the high \acp{dof} search space, have shown promise. Other techniques include inverse kinematics branching for iterative optimization of base and joint motions~\cite{bodily2017motion} and holistic control of the arm and base as a unified structure~\cite{haviland2022holistic}. 
While existing robotic planning methods achieve promising results on isolated tasks, such as door opening or object retrieval in controlled environments, their specialized, task-specific designs cannot generalize to broader scenarios requiring coordinated manipulation by the mobile base, manipulator arm, and target object. Yet, our \ac{smmp} scenario demands manipulation of objects with diverse kinematic structures in varied environments, where successful task execution critically depends on coordination among the mobile base, manipulator arm, and target object. 
In addition, deep \ac{rl} has recently gained popularity for manipulation tasks involving rich interactions. For example,~\cite{wang2020learningmobile} trains an \ac{rl} policy for object retrieval on a physical manipulator, and~\cite{ito2022efficient} uses deep \ac{rl} for whole-body control in door-opening tasks, while~\cite{xia2021relmogen} abstracts the action space into base and arm sub-goals for long-horizon tasks in simulated environments. Although \ac{rl} offers advantages for complex interaction planning, learned policies often suffer from poor transferability from simulation to real-world applications and do not scale effectively to long-horizon tasks due to substantial training time.

\subsection{\texorpdfstring{\ac{mmmp}}{}}
In sequential manipulation tasks, robots must repeatedly establish and release contact with various objects, exhibiting multi-modal behavior: contact states (discrete modes) constrain robot motions, effectively partitioning the environment's configuration space into interconnected manifolds. Transitions between manifolds indicate potential mode changes. Building on this concept, \ac{mmmp} methods~\cite{alami1994two,cambon2009hybrid,hauser2011randomized,barry2013hierarchical,toussaint2018differentiable} aim to find feasible trajectories across different manifolds, producing motion plans applicable to sequential mobile manipulation tasks. For example, Hauser~\etal~\cite{hauser2011randomized} propose a scalable algorithm that randomly samples mode switches and motion paths on a known mode transition graph to generate a solution plan, while Toussaint~\etal~\cite{toussaint2018differentiable} abstracts contact modes using differentiable physics, enabling tool-use planning. These methods yield impressive results in planning multi-step actions, but they share a limitation common to \ac{mmmp}: their planning domains are specifically designed and restricted to geometric features, necessitating extensive efforts in custom planner design and mode transition definitions for numerous contact modes. This proves inadequate for semantically rich environments where object relationships transcend simple contacts~\cite{garrett2020integrated}. While the \ac{mmmp} approach shares similarities with our work in terms of abstracting actions through contacts, the proposed \ac{akr} enables the use of off-the-shelf planning languages and aims to accommodate a broad range of mobile manipulation tasks without defining specific actions for each task.

\subsection{\texorpdfstring{\ac{tamp}}{}}
Thanks to the development of \ac{pddl}~\cite{mcdermott1998pddl} and other planning languages, complex symbolic planning can be solved using standard algorithms~\cite{helmert2006fast,karpas2020automated}. While symbolic planning effectively captures abstract concepts, it struggles to represent the feasibility of robot motions. This limitation has led the robotics community to integrate \ac{mmmp} concepts with symbolic task planners, forming the field of \ac{tamp}~\cite{garrett2020integrated}. Current \ac{tamp} approaches typically employ a bidirectional interface between task and motion planning~\cite{erdem2011combining,kaelbling2011hierarchical,srivastava2014combined,garrett2018ffrob}, but they remain computationally expensive due to their reliance on dense sampling in high-dimensional spaces~\cite{marcucci2023motion}. 
Recent work has sought to address these inefficiencies: Zhang~\etal~\cite{zhang2023symbolic} optimize symbolic state spaces to reduce redundant navigation actions, Yang~\etal~\cite{yang2024guiding} leverage \acp{vlm} to propose high-level subgoals to prune search spaces, and Sung~\etal~\cite{sung2023learning} learns back-jump heuristics that identify the culprit action and bypass irrelevant backtracking steps. 
However, the iterative nature of \ac{tamp} approaches still imposes significant computational overhead, as failed motion planning attempts trigger backtracking and replanning of action sequences. As a result, many \ac{tamp} approaches simplify motion planning and limit themselves to basic manipulation tasks, avoiding the complexity of designing intricate planning domains and specific motion planners for complex mobile manipulation tasks. 

Departing from traditional efforts in \ac{tamp} approaches that either optimize search strategies or redesign task-motion interfaces, this work proposes a new perspective by planning mobile manipulation tasks through the \ac{akr}. 
The proposed \ac{akr} constitutes an effective intermediate representation that can benefit \ac{tamp} in solving challenging sequential mobile manipulation tasks by improving computational efficiency through reducing intermediate variables and facilitating optimization-based motion planning.

This work builds upon our preliminary results presented in Jiao~\etal~\cite{jiao2021consolidated,jiao2021efficient}. The extension features a more comprehensive literature review, the introduction of a new plan refinement algorithm that enhances planning success rates by selecting key \ac{akr} configurations throughout the action sequence, and extensive benchmarking that compares our \ac{smmp} framework to baselines using off-the-shelf motion planners. Additionally, we include implementation and large-scale experimentation on a physical mobile manipulator platform.

\section{AKR Modeling}\label{sec:modeling}
This section describes three key steps for integrating robot and scene models into one cohesive kinematic representation, termed \acf{akr}. 

\subsection{Kinematic Representation}\label{sec:modeling:representation}
The kinematic representation used in this article is defined as a tree $\mathcal{T}=(V, S)$ where the rigid bodies of an articulated object are described as links $v_i\in V$, while their inherent motion constraints and spatial relations are represented by joints $s_{ij}\in S$. Specifically, the node set $V$ includes a set of links $v_i = \langle o_i, \mathcal{F}_i\rangle$; each encodes a full geometry model $o_i$ (\eg, a triangular mesh or empty for dummy link), and a link frame $\mathcal{F}_i$. In addition, the root node of $\mathcal{T}$ is denoted as $v_r$. The edge set $S$ includes a set of joints $s_{ij} = \langle r_{ij}, {}^{i}_{j}T \rangle$; each encodes the motion constraints $r_{ij}$ (\eg, bounded revolute or prismatic motion along an axis) between the parent link $v_i$ and the child link $v_j$, and a spatial transformation ${}^{i}_{j}T$ from the parent link frame $\mathcal{F}_i$ to the child link frame $\mathcal{F}_{j}$. Based on the above notations, a kinematic chain $\mathcal{C}_{ij}=(V_{ij}, S_{ij})$ contains only nodes $V_{ij}\subseteq V$ and edges $S_{ij}\subseteq S$ that belong to a path between a node $v_i$ and one of its descendant nodes $v_j$ in $\mathcal{T}$. 

\subsection{\texorpdfstring{\ac{akr}}{} Modeling for Mobile Manipulation}\label{sec:modeling:manipulation}
\begin{figure}[!t]
    \centering
    \includegraphics[width=0.9\linewidth]{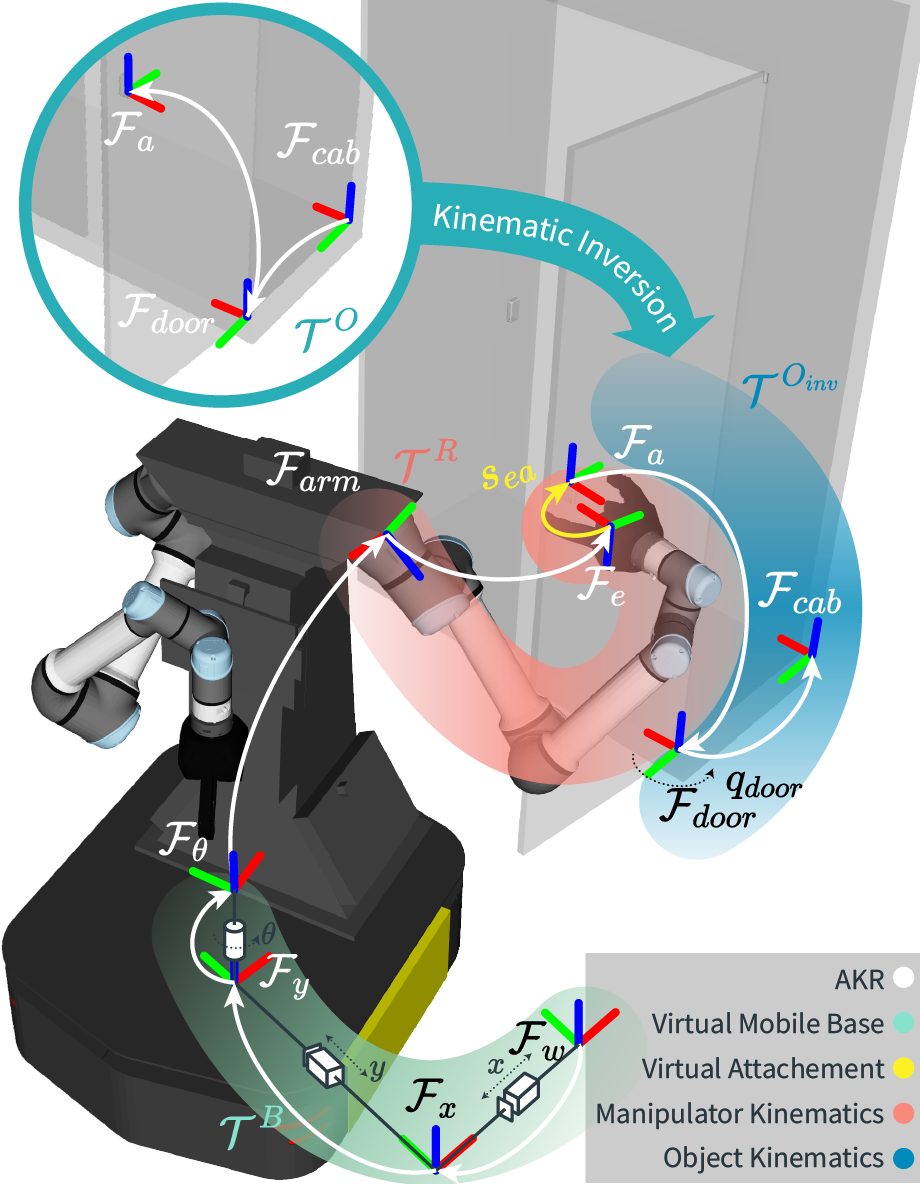}
    \caption{\textbf{Modeling a mobile manipulation task from the proposed \ac{smmp} perspective.} Constructing an \ac{akr} involves four key inputs: the manipulator's kinematics $\mathcal{T}^R$, object kinematics $\mathcal{T}^O$, a virtual mobile base $\mathcal{T}^B$, and a virtual attachment joint $s_{ea}$. Given the articulated nature of the drawer, its kinematic model requires inversion to maintain a tree.}
    \label{fig:akr}
    \vspace{-12pt}
\end{figure}

To construct an \ac{akr}, $\mathcal{T}^A$, four key inputs are required: the manipulator's kinematics $\mathcal{T}^R$, object kinematics $\mathcal{T}^O$, a virtual mobile base $\mathcal{T}^B$, and a virtual attachment joint $s_{ea}$ between the robot end-effector and the link to be grasping on the object. \cref{fig:akr} illustrates the constructed \ac{akr} for opening a cabinet door with a physical mobile manipulator platform. 

\textbf{\ac{akr} modeling} for a mobile manipulation planning problem first involves integrating the virtual mobile base into the kinematic model of the manipulator. During interactions, the post-inversion object's kinematics is further integrated into the \ac{akr} through a virtual attachment joint. The \ac{akr} is extended to the inverted object model's terminal link while maintaining its serial chain structure (see \cref{fig:akr}). In our application, we assume that $\mathcal{T}^R$ and $\mathcal{T}^O$ are known. However, obtaining the virtual mechanisms and constructing $\mathcal{T}^A$ involve more nuanced operations. The following section will detail these operations.

\textbf{Virtual mobile base} $\mathcal{T}^B$ reflects the motion possibilities of the mobile base. In \cref{fig:akr}, the manipulator with an omnidirectional mobile base can theoretically achieve free, stable motion on the ground plane. Consequently, a kinematic chain with three consecutive joints (two perpendicular prismatic joints are connected in serial to imitate linear motion, followed by one revolute joint at the rotation center of the mobile base to imitate angular motion) is sufficient to describe the motion of the base. 

\begin{algorithm}[t!]
\fontsize{8pt}{8pt}\selectfont
    \caption{Kinematics Inversion} 
    \label{alg:kin_inv}
    \LinesNumbered
    \SetKwInOut{KIN}{Input}
    \SetKwInOut{KOUT}{Output}
    \SetKwInOut{Param}{Params}
    \KIN{The kinematics: $\mathcal{T}=(V,S)$,\\
         The root node of $\mathcal{T}$: $v_r$,\\
         The attachable link node: $v_a$.\\ (not necessarily the terminal node)
         }
    \KOUT{The inverted kinematics: $\mathcal{T}^\text{inv}=(V,S^\text{inv})$.}
    \textcolor{blue}{\tcp*[h]{Initialization}}\\
    $S^\text{inv}\gets\{\}$\;
    \textcolor{blue}{\tcp*[h]{Get kinematic chain from $v_r$ to $v_a$}}\\
    $(V_{ra},S_{ra})\gets FindPath(\mathcal{T},v_r,v_a)$\;
    \textcolor{blue}{\tcp*[h]{Inversion of the kinematic chain}}\\
    \ForEach {$\{s_{ij}, s_{jk}\} \subset S_{ra}$}
    {
        $(r_{ij}, {}^{i}_{j}T)\gets s_{ij}$, $(r_{jk}, {}^{j}_{k}T)\gets s_{jk}$\;
        \eIf{$v_k$ is equal to $v_a$}
        {   
            $s^*_{ji}\gets (r_{ji}, {}^{j}_{k}T^{-1})$\;
            $s^*_{kj}\gets (r_{kj}, I_4)$\;
            $S^\text{inv}\gets S^\text{inv}\cup\{s^*_{ji},s^*_{kj}\}$\;
        }
        {   
            $s^*_{ji}\gets (r_{ji}, {}^{j}_{k}T^{-1})$\;
            $S^\text{inv}\gets S^\text{inv}\cup\{s^*_{ji}\}$\;
        }
    }
    \textcolor{blue}{\tcp*[h]{Inversion of branches}}\\
    \ForEach {$v_j \in V_{ra}$ and $v_j \neq v_r$}
    {
        \ForEach{$s_{jk}\in S$ and $s_{jk}\not\in S_{ra}$}
        {
            $(r_{jk}, {}^{j}_{k}T)\gets s_{jk}$\;
            $(r_{ij}, {}^{i}_{j}T)\gets s_{ij},\ \text{where}\ s_{ij}\in S^\text{inv}$\;
            $s^*_{jk} \gets (r_{jk},{}^{i}_{j}{T}{}^{j}_{k}{T})$\;
            $S^\text{inv}\gets S^\text{inv}\cup\{s^*_{jk}\}$\;
        }
    }
    \ForEach {$v_i \in V$ and $v_i \not\in V_{ra}$}
    {
        \If{$\exists s_{ij}\in S$}{
            $S^\text{inv}\gets S^\text{inv}\cup\{s_{ij}\}$\;
        }
    }
    \textcolor{blue}{\tcp*[h]{Get the inverted kinematics}}\\
    $\mathcal{T}^\text{inv}\gets(V,S^\text{inv})$\;
\end{algorithm}

\textbf{Virtual attachment joint} $s_{ea}$ characterizes the motion constraints and spatial relation between the robot and the scene after integration. As shown in \cref{fig:akr}, by inserting the $s_{ea}$ between the manipulator's end-effector link $v_e$ and an attachable link $v_a$ in the object model, the kinematics of the mobile manipulator and the manipulated object are integrated. If a manipulated object $\mathcal{T}^O$ is articulated and the attachable link is not the root node of $\mathcal{T}^O$, its kinematic model must be inverted before integration to ensure that $\mathcal{T}^A$ remains a tree (\ie, each node within a tree has at most one parent node).

\textbf{Kinematics inversion} process reverses the kinematic model of the manipulated object while retaining its motion constraints and geometric consistencies, as shown in \cref{alg:kin_inv}. 
Our kinematic tree representation defines transformations from parent to child link frames, with motion constraints (\ie, joints) specified relative to the child frame. Therefore, kinematics inversion requires non-trivial adjustments to each joint's spatial transformation, in addition to simple parent-child inversions, since joints constrain child link motion relative to their local frame. The algorithm first identifies the main branch (Line 4) between the base link (the root node of $\mathcal{T}$) and the attachable link (identified in $s_{ea}$), including all intermediate joints and nodes. Transformations along this kinematic chain (between \( v_r \) and \( v_a \)) are then updated (Lines 6-14). The motion planner treats side branches as static, but their proper geometric transformation (Lines 16-24) remains critical for maintaining self-collision avoidance in the \ac{akr} representation. \cref{fig:akr} illustrates the post-inversion cabinet kinematics and its integration into the \ac{akr}.

From the \ac{akr} perspective, we can formulate a \textit{single-step mobile manipulation} task as motion planning in A-Space (\ie, the configuration space of \ac{akr}), with task execution represented by \ac{akr} state transitions. Unlike decoupled approaches that treated the object as task-specific constraints imposed on the robot, \eg,~\cite{stilman2010global}, the \ac{akr} simultaneously incorporates: 1)~kinematic constraints for both robot and manipulated object, 2)~path constraints for end-effector during interaction, and 3)~self-collision avoidance\textemdash{}enabling generation of safe, coordinated base-arm-object motions. By generalizing to objects with known kinematics, the \ac{akr} eliminates task-specific modeling requirements for diverse objects and environments. This approach achieves effective whole-body motion optimization by eliminating the need for iterative backtracking in base-arm-object coordination at the task level.

\section{Planning in the A-Space}\label{sec:planning}
In this section, we first formulate motion planning problems for \textit{single-step} mobile manipulation tasks in A-Space, and solve them via warm-started trajectory optimization. Then, we tackle the \textit{multi-step} \ac{smmp} problem through an \ac{akr}-based task planning, supporting the generation of whole-body trajectories for interacting with multiple objects sequentially, implemented via three action predicates.

\subsection{Motion Planning in A-Space}\label{sec:planning:motion}
Consider a standard single-arm mobile manipulation task, in which a mobile manipulator interacts with an articulated object within the scene. The state vector $\pmb{q}^\top=[\pmb{q}^{B},\pmb{q}^{R},\pmb{q}^{O}]^\top\in\mathcal{Q}^{\text{free}}$ describes the state of the virtual mobile base $\mathcal{T}^B$, the manipulator $\mathcal{T}^R$, and the articulated object $\mathcal{T}^O$, respectively. Notably, these joints belong to a serial kinematic chain $\mathcal{C}$, which consists of a root node $v_w$ and a non-root node $v_b$, as illustrated in \cref{fig:akr}. The remaining joints that do not belong to $\mathcal{C}$ are assumed to be fixed during motion planning. $\mathcal{Q}^{\text{free}}\subset\mathbb{R}^n$ is the collision-free subset of A-Space. The motion planning problem in A-Space is equivalent to finding a $T$-step path $\pmb{q}_{1:T}=\langle \pmb{q}_1,\pmb{q}_2,\ldots,\pmb{q}_T \rangle\in \mathcal{Q}^{\text{free}}$, which can be formulated and solved by trajectory optimization. 

Following Jiao~\etal~\cite{jiao2021consolidated}, the trajectory optimization problem is formulated as:
\begin{align}
    \underset{\pmb{q}_{1:T}}{\text{minimize}} &\sum_{t=1}^{T-1} ||\pmb{W}_{v}\delta{\pmb{q}}_{t} ||_2^2
    \; + \sum_{t=2}^{T-1} || \pmb{W}_{a}\delta\dot{\pmb{q}}_{t} ||_2^2,\label{eqn:opt_objective}\\ 
    &h_{\text{chain}}(\pmb{q}_t) = 0, \; \forall t = 1, 2, \ldots, T,\label{eqn:opt_chain}\\ 
    &|| f_{\text{task}}(\pmb{q}_T) - \pmb{g}_{\text{goal}} ||^2_2 - \xi_{\text{goal}} \leq 0,\label{eqn:opt_goal}
\end{align}
where \cref{eqn:opt_objective} penalizes the overall traveled distance and overall non-smoothness of the trajectory $\pmb{q}_{1:T}$. $\pmb{W}_{v}$ and $\pmb{W}_{a}$ are two diagonal weighting matrices for each \ac{dof}, $\delta{\pmb{q}}_{t}$ and $\delta\dot{\pmb{q}}_{t}$ are the finite forward difference and second-order finite central difference of $\pmb{q}_{t}$, respectively. 
The equality constraint \cref{eqn:opt_chain} specifies the physical constraints of the object or the environment during interactions. Failing to account for this type of constraint (\eg, the kinematic constraint of the robot and the scene) may damage the robot or the manipulated object, resulting in failed executions. 
The goal of a mobile manipulation task is bounded through an inequality constraint \cref{eqn:opt_goal} with a tolerance $\xi_{\text{goal}}$. The function $f_{\text{task}}: \mathbb{R}^n\rightarrow\mathbb{R}^k$ maps $\pmb{q}_T$ from the configuration space $\mathcal{Q}$ to the task-dependent goal space $\mathcal{G}\in\mathbb{R}^k$. For instance, in an object-picking task, $f_{\text{task}}$ represents the forward kinematics used to compute the robot's end-effector pose, while $\pmb{g}_{\text{goal}}$ denotes the end-effector goal pose before grasping. In a door-opening task, $f_{\text{task}}$ maps the \ac{akr} state to the door's joint configuration, and $\pmb{g}_{\text{goal}}$ represents the desired joint angle for the door.

Additional safety constraints are imposed during trajectory optimization.
Without loss of generality, we assume an omnidirectional base and only kinematic constraints in this paper. However, additional constraints, such as nonholonomic constraints for non-omnidirectional mobile bases could be formulated into the optimization problem by incorporating additional terms~\cite{rosmann2017kinodynamic}:
\begin{align}
    \pmb{q}_{\text{min}} \leq \pmb{q}_t &\leq \pmb{q}_{\text{max}}, \, \;\;\;\;\; \forall t = 1, 2, \ldots, T%
    \label{eqn:cnt_jnt}%
    \\
    ||\delta{\pmb{q}}_{t}||_\infty &\leq \delta{\pmb{q}}_{\text{max}}, \, \;\;\; \forall t = 1, 2, \ldots, T-1%
    \label{eqn:cnt_vel}%
    \\
    ||\delta\dot{\pmb{q}}_{t}||_\infty &\leq \delta\dot{\pmb{q}}_{\text{max}}, \, \;\;\; \forall t = 2, 3, \ldots, T-1%
    \label{eqn:cnt_acc}%
    \\
    \sum_{i= 1}^{N_{\text{link}}} \sum_{j=1}^{N_{\text{obj}}} &|\text{dist}_{\text{safe}} - sd(L_i, O_j)|^{+} \leq \xi_{\text{dist}}%
    \label{eqn:collision_1},%
    \\
    \sum_{i= 1}^{N_{\text{link}}} \sum_{j=1}^{N_{\text{link}}} &|\text{dist}_{\text{safe}} - sd(L_i, L_j)|^{+} \leq \xi_{\text{dist}}%
    \label{eqn:collision_2},%
\end{align}
where $|\cdot|^{+}$ is defined as $|x|^{+}=\text{max}(x,0)$.
\cref{eqn:cnt_jnt,eqn:cnt_vel,eqn:cnt_acc} are inequality constraints that define the joint capability and implicitly constrain the workspace of both the robot and the scene. \cref{eqn:collision_1} and \cref{eqn:collision_2} penalize collisions with obstacles and self-collisions, respectively. $N_{\text{link}}$ and $N_{\text{obj}}$ are the number of links that belong to the \ac{akr} and the number of obstacle objects within the scene, respectively. $\text{dist}_{\text{safe}}$ is a predefined safety distance, and $sd(\cdot)$ is a function that calculates the signed distance between a pair of objects. $\xi_{\text{dist}}$ is a collision tolerance parameter. The formulated problem is solved through trajectory optimization~\cite{schulman2014motion}.

Unlike sampling-based methods, optimization-based motion generation methods rely on gradient descent algorithms and can easily become trapped in undesired local minima near the initial guess~\cite{ratliff2009chomp,schulman2014motion}. Consequently, a proper trajectory initialization (\ie, warm start) is essential to improve the optimization results. However, the high dimensionality of \ac{akr} presents significant challenges. While sampling-based methods can provide paths as initialization seeds for optimization, they become computationally expensive in high-dimensional \ac{akr} spaces. Simple interpolation between start and goal states is insufficient, as base movements are often constrained by cluttered obstacles. Solving for coordinated movements simultaneously creates a complex optimization landscape with many poor local minima, making convergence difficult without good initialization. Therefore, an efficient initialization strategy that balances computational cost with solution quality is crucial for making \ac{akr}-based planning practical.

Therefore, we devise an A$^\star$-based trajectory initialization method to effectively guide trajectory optimization away from poor local minima without requiring excessive computational time. Given the initial state $\pmb{q}_i$ and the goal state $\pmb{q}_g$, this method utilizes A$^\star$ to find a feasible path from the current location $\pmb{q}_i^B$ to the goal location $\pmb{q}_g^B$ of the mobile base. Subsequently, it linearly interpolates the manipulator's joint state from $\pmb{q}_i^R$ to $\pmb{q}_g^R$. 
While the method itself is presented as a simple design, it is pivotal to our framework's practical efficacy. The initialization phase aims to generate coarse, collision-free base paths (via $A^\star$) to guide subsequent trajectory optimization. Without this step, the solver often converges to local minima—for example, favoring shorter but colliding base paths over safer, longer ones.  \cref{sec:supp:trajinit} provides quantitative comparisons with baselines, effectiveness analysis, and discussions of trade-offs and limitations.

\subsection{Task Planning for Sequential Tasks}\label{sec:planning:task}
To solve a \ac{smmp} problem, a robot must break it down into a sequence of temporally feasible actions, necessitating task planning. Following the classic formalization of task planning, we describe the environment by a set of states $\mathcal{E}$ (of note, $\mathcal{E}$ and $\mathcal{Q}$ are unnecessarily identical). Possible transitions between these states are defined by $\mathcal{A}\subseteq\mathcal{E}\times\mathcal{E}$, where a transition $a=\langle e,e' \rangle \in\mathcal{A}$ alters the environment state from $e\in\mathcal{E}$ to $e'\in\mathcal{E}$. The task planning goal is to identify a sequence of transitions $a_{1:N}$ that alter the environment from its initial state $e_0\in \mathcal{E}$ to a goal state $e_N\in \mathcal{E}_g$, where $\mathcal{E}_g \subseteq \mathcal{E}$ is a set of goal states. Traditional task planning involves defining meaningful symbolic actions $\mathcal{A}$ and states $\mathcal{E}$ and often assumes a robot can execute the elementary actions. However, these symbolic actions necessitate substantial manual design effort to be instantiated successfully at the motion level. From the \ac{akr} perspective, the action is defined as changes to the \ac{akr} structure and corresponding A-Space, transforming the \ac{smmp} problem into a series of \ac{akr} structural modifications.

In this section, we describe how \texttt{actions} defined using standard planning language (\eg, \ac{pddl}~\cite{mcdermott1998pddl}) can be used to properly formulate a task planning problem, and how the planned actions sequence can be realized by motion planning within A-Space. We start by making connections between the action semantics and the actual manipulation behaviors, before explaining how motion planners process the predicates and variables in the action definitions.

\paragraph*{\texttt{goto-akr} (\texttt{akr}, $\pmb{q_1}$, $\pmb{q_2}$)}
This predicate moves the A-Space state from pose $\pmb{q}_1$ to the desired pose $\pmb{q}_2$. It represents the tasks that do not require interaction with the environment, wherein the \ac{akr} structure remains unchanged. Pure navigation is a typical action that falls into this category. 

\paragraph*{\texttt{pick-akr} (\texttt{akr}, \texttt{o}, \texttt{s})}
This predicate moves the \ac{akr} to an object, \texttt{o}, with kinematics $\mathcal{T}$ and extends the current \ac{akr}'s kinematics, by adding a virtual attachment joint \texttt{s}$\gets s_{ea}$ to connect the object and the arm's end-effector. In practice, $s_{ea}$ encodes both the end-effector's grasping pose and the associated grasp constraints between the robot and the object. \texttt{pick-akr} represents the group of tasks that require mobile manipulators to interact with the environment, \eg, picking up an object or grasping a handle.

\paragraph*{\texttt{place-akr} (\texttt{akr}, \texttt{o}, \texttt{g})}

This predicate moves the object, \texttt{o}, connected to \texttt{akr} to an object-specific goal state \texttt{g}, while the object to be manipulated is incorporated into the \ac{akr} and imposes kinematic constraints. For example, \texttt{g} represents the target door state (\eg, \texttt{opened}) in door-opening tasks or the desired object placement location (\eg, \texttt{onTable}) in object relocation tasks. Once the goal state is reached, \texttt{place-akr} breaks the current \ac{akr} at the virtual attachment joint where it connects the mobile manipulator and the object, and the object will be placed where it was disconnected from the \ac{akr}. \texttt{place-akr} represents the group of tasks for which mobile manipulators stop interacting with the environment, such as placing an object on the table.

The primary challenge in generalizing actions across objects stems from heterogeneous task-specific constraints tied to scenes and objects, which is pivotal for generating executable trajectories in different mobile manipulation tasks. By embedding these constraints into scene kinematics, the \ac{akr} achieves a unified action definition and enables a general formulation of trajectory optimization for both rigid and articulated objects with known kinematics. This \ac{akr}-based formulation alleviates the need to define task-specific actions for manipulating different objects, which in turn reduces the need for intermediate subgoals (\eg, moving the mobile base near the object before manipulation) and allows more dexterous exploration of the A-Space.

\section{Sequential Mobile Manipulation Planning}\label{sec:refine}
In this section, we first present the operation of the plan refinement algorithm. We then describe how the algorithm resolves motion infeasibility in sequential tasks by selecting favorable action parameters through a goal selection process.

\subsection{Plan Refinement for Sequential Tasks}\label{sec:refine:example}
\begin{figure}[t!]
    \centering
    \includegraphics[width=\linewidth]{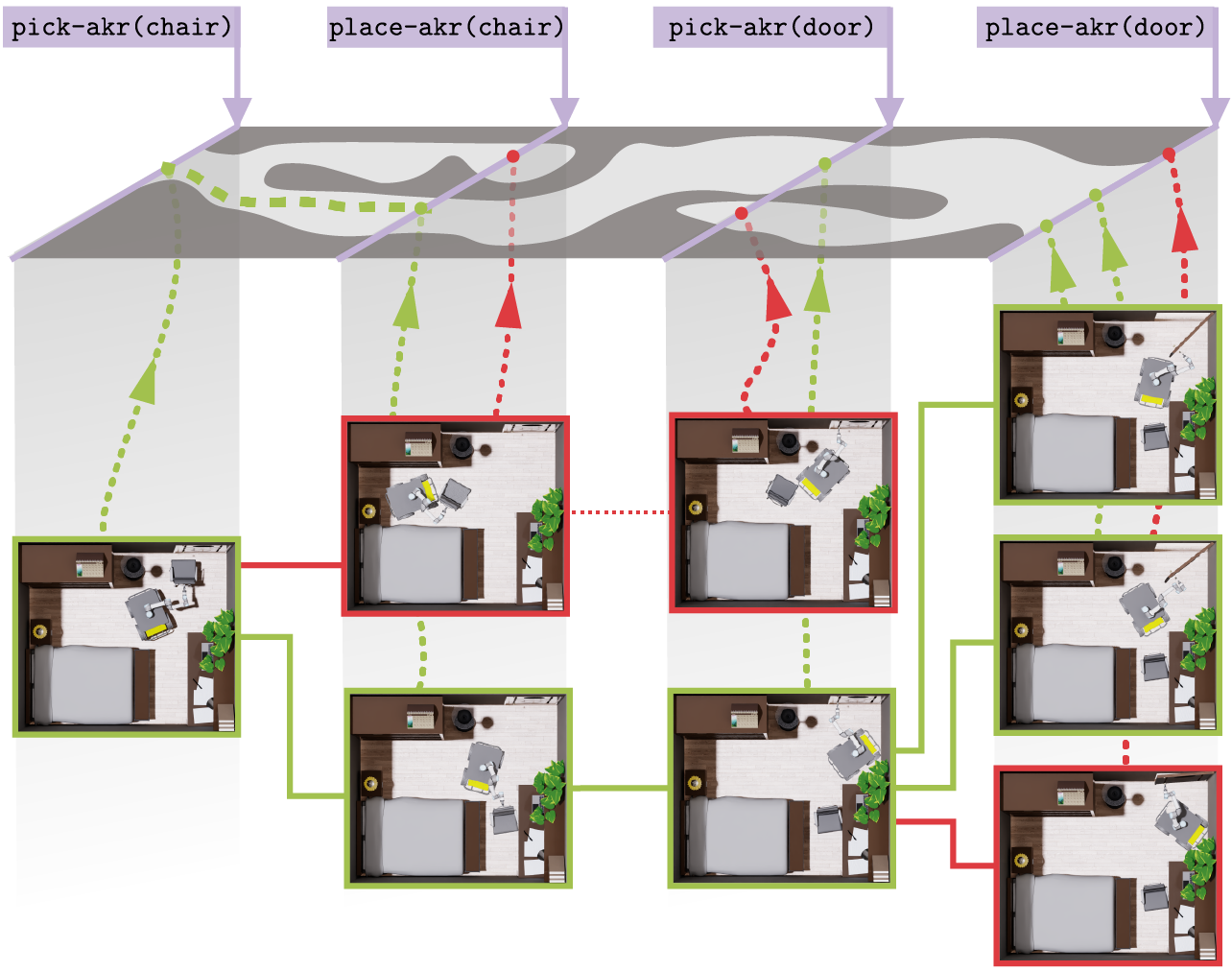}
    \caption{\textbf{An illustration of the proposed plan refinement algorithm for a sequential task.} The task planner first generates a sequence of symbolic actions representing the evolution of the \ac{akr} configuration space, indicating robot-environment interactions. Then, the plan refinement algorithm ensures motion feasibility among sampled \ac{akr} end configurations for consecutive actions.}
    \label{fig:smmp}
\end{figure}

To illustrate how symbolic action predicates (as defined in \cref{sec:planning:task}) govern the evolution of the \ac{akr} structure and how plan refinement resolves motion feasibility, consider the example in \cref{fig:overview}(a). The robot must first relocate a chair blocking a door (T1) and then open the door to exit (T2), interacting with two articulated objects. The action sequence includes four steps as shown in \cref{fig:smmp}.

The \ac{akr} evolving with each action depicts possible end states for each step. The first action, \texttt{pick-akr}, generates a whole-body motion for the \texttt{akr} (virtual mobile base and manipulator) to grasp the chair. After grasping, \texttt{akr} integrates the chair's kinematics into a new \texttt{akr}. The resulting A-Space captures the kinematics of the mobile base, manipulator, and chair, while enforcing a planar constraint on the new \texttt{akr}'s end-effector (\ie, the chair's base link) to emulate the chair’s planar motion across the floor. These constraints are collectively considered during trajectory optimization.

For \texttt{place-akr}, the robot must choose a chair placement that avoids blocking subsequent door access. We sample valid configurations within A-Space and illustrate two representative configurations (second column in \cref{fig:smmp}(c)). Without considering future actions, both placements are acceptable, as the chair no longer obstructs the door. After placing the chair, the \texttt{akr} detaches it, reverting to the mobile manipulator. The subsequent \texttt{pick-akr} action finds the path to approach the door handle. By sampling valid configurations (third column in \cref{fig:smmp}), we can distinguish between motion-infeasible (red) and motion-feasible (green) pairs. Due to the presence of motion-infeasible pairs, plan refinement becomes necessary. 

The plan refinement process acts as a receding horizon, evaluating feasibility across the action sequence. For instance, the last configuration in \cref{fig:smmp}(c) (fourth column) is infeasible due to the subsequent action that approaches the drawer, which requires the robot to pass through the door. This example demonstrates how previously defined actions govern the evolution of the \ac{akr} structure and underscores the necessity of plan refinement in resolving motion feasibility. The subsequent section formalizes this procedure, focusing specifically on the selection of action parameters.

\subsection{Action Parameter Selection from Key Configuration Set}
The \ac{akr}-based motion planner requires two sets of action parameters to generate trajectories. The first, end-effector poses \texttt{s}, specifies grasps between the robot's end-effector and objects during \texttt{pick-akr} actions. These poses are obtained through grasp synthesis methods (\eg, \cite{morlans2024grasp}) or predefined for known objects, as grasp generation lies beyond the scope of this work. The second set, \texttt{g}, defines object-centric states aligned with symbolic predicates, such as an \texttt{opened} door or an object placement goal like \texttt{onTable}, and must be instantiated appropriately within the object's configuration space for trajectory optimization~\cite{garrett2020pddlstream}.

Improper action parameters values can lead to motion infeasibility, as they do not fully capture the state of the \texttt{akr}, and variations in \texttt{akr} states impose different feasibility conditions on subsequent actions, as illustrated in \cref{fig:smmp}. While multiple chair placements are feasible in a bedroom, some configurations (marked red) obstruct the robot from approaching the door due to self-blocking. This challenge is exacerbated in confined spaces with limited configuration space. To ensure feasibility across action sequences, our method jointly optimizes goal states with future steps, resolving conflicts during plan refinement to avoid such pitfalls.

The exhaustive motion planning for all possible action parameters is computationally demanding, with time complexity growing exponentially with action sequence length, rendering it impractical. To address this challenge, we propose a plan refinement algorithm designed to efficiently select the goal \ac{akr} state by considering a given number of anticipated subsequent actions. This approach aims to improve the likelihood of success for sequential tasks.

Specifically, let $\pmb{q}_{a_n}$ be a possible goal \ac{akr} configuration for the action $a_n$ and $\mathcal{Q}_{a_n}$ be the A-Space during that action, and $\mathcal{Q}_{a_{n:n+l}}$ be the Cartesian product of A-Spaces: $\mathcal{Q}_{a_{n:n+l}} = \mathcal{Q}_{a_n} \times \mathcal{Q}_{a_{n+1}} \times \ldots \times \mathcal{Q}_{a_{n+l}}$, where $l$ is the window length suggesting the number ($l+1$) of anticipated subsequent actions. Our aim is to find a \ac{kcs} $ \pmb{q}^*_{a_{n:n+l}} = \langle \pmb{q}_{a_n}, \pmb{q}_{a_{n+1}}, \ldots,  \pmb{q}_{a_{n+l}} \rangle \in \mathcal{Q}_{a_{n:n+l}} $ so that transition among every two consecutive configurations is valid and efficient. 

\begin{algorithm}[t!]
\fontsize{8pt}{8pt}\selectfont
    \caption[]{Select \ac{kcs}}
    \label{alg:keyset}
    \LinesNumbered
    \SetKwInOut{KIN}{Input}
    \SetKwInOut{KOUT}{Ouput}
    \SetKwInOut{PARAM}{Params}

    \KIN{Action sequence segment: $a_{n:n+l}$\\
    Current \ac{akr}: $\mathcal{T}^A_{a_{n-1}}$\\
    Current \ac{akr} state: $\pmb{q}_{a_{n-1}}$\\
    }

    \KOUT{Preferred key configuration set: $\pmb{q}^*_{a_{n:n+l}}$
    }

    \PARAM{No. of candidate configurations: $N_c$ \\
           No. of clusters: $N_k$ \\
           No. of anticipated subsequent actions: $l$ \\
    }

    $\pmb{q}_{a_{n-1:n-1}} \gets \langle \pmb{q}_{a_{n-1}} \rangle $
    
    $K \gets \{ \pmb{q}_{a_{n-1:n-1}} \}$
    
    \For{\(i \in n:n+l\)}{

    $K_{temp} \gets \emptyset$

    \textcolor{blue}{\tcp*[h]{Update A-Space according to \cref{sec:planning:task}.}}\\
    $\mathcal{T}^A_{a_i} \gets ConstructAKR(\mathcal{T}^A_{a_{i-1}}, a_i)$\\
    \textcolor{blue}{\tcp*[h]{Generate valid configurations within A-Space.}}\\
    $Q_{a_i} \gets SampleValidConfigurations(\mathcal{T}^A_{a_i}, N_c)$\\
    \textcolor{blue}{\tcp*[h]{Pruning similar configurations through down-sampling.}}\\
    $Q'_{a_i} \gets Downsample(Q_{a_i}, N_k)$\\
    \textcolor{blue}{\tcp*[h]{Predict and store feasible \ac{kcs}}}\\
    \For{$\pmb{q}_{a_{n-1:i-1}} \in K$}{
    
    \For{$\pmb{q}_{a_i} \in Q'_{a_i}$}{
    
    \If{$CheckMotionFeasibility(\mathcal{T}^A_{a_i}, \pmb{q}_{a_{n-1:i-1}}, \pmb{q}_{a_i})$}{

    $\pmb{q}_{a_{n-1:i}} \gets \pmb{q}_{a_{n-1:i-1}}.append(\pmb{q}_{a_i}) $
    
    $K_{temp} \gets K_{temp} \cup \{ \pmb{q}_{a_{n-1:i}}  \} $
    
    }}}    
    $K \gets K_{temp}$
    }

    \textcolor{blue}{\tcp*[h]{Select \ac{kcs} of lowest cost}}\\
    $\pmb{q}^*_{a_{n-1:n+l}} \gets SelectBest(K) $

    $\pmb{q}^*_{a_{n:n+l}} \gets \pmb{q}^*_{a_{n-1:n+l}} \setminus \{ \pmb{q}_{a_{n-1}} \} $

\end{algorithm}

\cref{alg:keyset} details the process. The algorithm takes three inputs: 1)~a segment of the action sequence \(a_{n:n+l}\), 2)~the current \ac{akr} structure \(\mathcal{T}^A_{a_{n-1}}\), and 3)~the current \ac{akr} state \(\pmb{q}_{a_{n-1}}\). Parameters include \(N_c\), the number of candidate configurations sampled per action; \(N_k\), the number of clusters for down-sampling; and \(l\), the horizon length for anticipated actions. The subsequent paragraphs detail phases of the workflow.

\textbf{A-Space construction (line 6):} For each action \(a_i\) in the sequence \(a_{n:n+l}\), the algorithm first updates the \ac{akr} structure \(\mathcal{T}^A_{a_i}\) by integrating the kinematics of the manipulated object (\eg, a door or chair) into the robot's kinematics, as detailed in \cref{sec:modeling:manipulation}. This constructs the A-Space, which encodes the combined configuration space of the robot and object.

\textbf{Configuration sampling (line 8):} We define $Q_{a_i}\subset\mathcal{Q}_{a_i}$ as the finite set of sampled configurations for \ac{akr} $\mathcal{T}^A_{a_{i-1}}$, where each $\pmb{q}_{a_i} \in Q_{a_i}$ satisfies task-specific goal constraints and collision-free conditions. To fully explore possible goal configurations, we formulate an optimization problem:
\begin{align}
    \min_{\pmb{q}_{a_i}}\quad{} & || h_{\text{chain}}(\pmb{q}_{a_i}) ||^2_2+|| f_{\text{task}}(\pmb{q}_{a_i}) - \pmb{g}_{a_i} ||^2_2 \label{eqn:supp:ckcs_objective} \\
    \textrm{s.t.} \quad{} & || h_{\text{chain}}(\pmb{q}_{a_i}) ||^2_2 \leq \xi_{\text{chain}} \label{eqn:supp:ckcs_chain_cnt}                      \\
                          & || f_{\text{task}}(\pmb{q}_{a_i}) - \pmb{g}_{a_i} ||^2_2 \leq \xi_{\text{goal}} \label{eqn:supp:ckcs_goal_cnt}               \\
                          & \pmb{q}^{\min} \leq \pmb{q}_{a_i} \leq \pmb{q}^{\max} \label{eqn:supp:ckcs_joint_limit}                                            \\
                          & \pmb{q}_{a_i}\in \mathcal{Q}_{a_i}^{\text{free}} \label{eqn:supp:ckcs_collision}
\end{align}
where \cref{eqn:supp:ckcs_objective} penalizes the violation of the environment and the goal constraint corresponding to \cref{eqn:opt_chain,eqn:opt_goal}, \cref{eqn:supp:ckcs_chain_cnt,eqn:supp:ckcs_goal_cnt} bound the objective with a small tolerance to reduce undesirable results, \cref{eqn:supp:ckcs_joint_limit} constrains the \(\pmb{q}_{a_i}\) to be within the joint limit of the \ac{akr}, including both the robot and the manipulated object, \cref{eqn:supp:ckcs_collision} ensures \(\pmb{q}_{a_i}\) is collision-free in the environment. \cref{alg:samplevalidconfig} details how to solve the problem to generate a finite \(Q_{a_i}\) for \(a_i\). We first randomly sample goal \ac{akr} configurations based on the constructed \ac{akr} \(\mathcal{T}^A_{a_i}\) from a uniform distribution to initialize \(computeIK\) and compute the inverse kinematics problem (\ie, \cref{eqn:supp:ckcs_objective}) numerically on \(\mathcal{T}^A_{a_i}\). Solutions satisfying \cref{eqn:supp:ckcs_chain_cnt,eqn:supp:ckcs_goal_cnt,eqn:supp:ckcs_joint_limit} are added to \(Q_{a_i}\) until reaching its maximum cardinality. Then, we prune out configurations that are in collisions (\ie, violating \cref{eqn:supp:ckcs_collision}). Note that the collision check is reserved until the last step because collisions frequently happen in a confined and cluttered environment, and checking collisions is computationally heavy.

\begin{algorithm}[t!]
\fontsize{8pt}{8pt}\selectfont
    \caption[]{SampleValidConfigurations}
    \label{alg:samplevalidconfig}
    \LinesNumbered
    \SetKwInOut{KIN}{Input}
    \SetKwInOut{KOUT}{Ouput}
    \SetKwInOut{PARAM}{Params}
    \KIN{An \ac{akr}: \(\mathcal{T}^A_{a_i}\)\\
    }
    \KOUT{The Set of Valid Configurations: \(Q_{a_i}\) \\
    }

    \PARAM{Max. Cardinality of \(Q_{a_i}\): \(|Q_{a_i}|^\textit{max}\)\\
        Max. Tries of IK Calculation: \text{MAX\_TRIES}\\
    }
    \(Q_{a_i} \gets \emptyset\)\\
    \(counts \gets 0\)\\
    \While{\(|Q_{a_i}|<|Q_{a_i}|^\text{max}\) or \(counts < \)\text{MAX\_TRIES}}{
    \(\pmb{q}_{a_i} \gets computeIK(\mathcal{T}^A_{a_i})\) //\wrt \cref{eqn:supp:ckcs_objective} \ \\
    \If{\(\pmb{q}_{a_i}\) satisfy \cref{eqn:supp:ckcs_chain_cnt,eqn:supp:ckcs_goal_cnt,eqn:supp:ckcs_joint_limit,eqn:supp:ckcs_collision}}{
        \(Q_{a_i} \gets Q_{a_i} \cup \{\pmb{q}_{a_i}\}\)
    }
    \(counts\)++
    }
\end{algorithm}

\textbf{Configuration down-sampling (line 10):} Even after discarding configurations that are in collision, the candidate set \(Q_{a_i}\) often remains large. This can make motion feasibility checking computationally expensive due to the combinatorial nature of validating transitions across a sequence of actions\textemdash{}requiring up to \( |Q_{a_n}| \times |Q_{a_{n+1}}| \times \ldots \times |Q_{a_{n+l}}| \) checks. While retaining all candidates helps preserve completeness, in practice, down-sampling the configuration set can significantly improve planning efficiency by reducing the number of costly, repetitive feasibility checks. Thus, we down-sample configurations based on the assumption that those located close to each other in configuration space (\ie, with similar joint values and small Euclidean distances) exhibit similar motion feasibility. This is justified by the nature of our \ac{akr}-based trajectory optimization: \cref{eqn:opt_goal} constrains only a subset of the \ac{akr} state variables via $f_\text{task}$. As a result, nearby configurations often converge to the same local minima during trajectory optimization, making it redundant to plan from each configuration individually.

To down-sample \(Q_{a_i}\) and avoid redundant computations for similar configurations, we use the k-means++ method~\cite{arthur2007k} to partition \(Q_{a_i}\) into \(N_k\) clusters by minimizing the variance within the cluster: \(Q_{a_i} = \{Q^1_{a_i}, \ldots, Q^k_{a_i}\}\) with the corresponding cluster centroid \(\bar{Q}^k_{a_i}\). Then we construct a downsampled set \(Q'_{a_i}=\{\pmb{q}'^{1}_{a_i}, \pmb{q}'^{2}_{a_i}, \ldots, \pmb{q}'^{k}_{a_i} \}\) by selecting \(\pmb{q}'^{k}_{a_i}\) that is closest to the centroid in each cluster as the key configuration for the whole action sequence. Note that the cluster centroid itself may not be a valid configuration.

We acknowledge that the down-sampling step introduces some incompleteness. However, we wish to clarify that down-sampling is primarily a practical strategy to improve efficiency, as further experiments in \cref{sec:res:ablation} demonstrate. While the current implementation focuses on empirical performance, we believe that completeness can be achieved through a more sophisticated and structured sampling strategy, as suggested by previous work~\cite{garrett2020pddlstream, zhang2023symbolic, sung2023learning}.

\textbf{Feasibility Checking (line 12-17):} As the above procedure produces a much more compact \(Q'_{a_{n:n+l}}\), checking the motion feasibility among its elements becomes feasible. Specifically, \(checkMotionFeasibility\) estimates the motion feasibility for \(\langle \pmb{q}_{a_i},\pmb{q}_{a_{i+1}} \rangle\) by applying the \(A^\star\) algorithm (the map and base path are reused for trajectory initialization to reduce computational effort) to find a path between the mobile base poses encoded in key configurations. We will record the key configuration in $K$ if there is a feasible base path. In the example shown in \cref{fig:smmp}(c), the key configuration in red is removed because no viable path connects it to the upcoming action of grasping the door handle or passing through the door.

\textbf{Optimal \ac{kcs} Selection (line 19):} The procedure iterates until all actions within horizon $l$ are checked, resulting in the construction of $K$, which consists of feasible \ac{kcs}. Subsequently, we employ an objective function to penalize the total traveling distance and select the best \ac{kcs} $\pmb{q}^*_{a_{n:n+l}}$ with minimal cost:
\begin{equation}
    \min_{\pmb{q}_{a_{n-1:n+l}}} \sum_{i=n}^{n+l} ( ||\pmb{W}_R(\pmb{q}^R_{i-1} - \pmb{q}^R_{i}) ||^2_2 + ||\pmb{W}_B (\pmb{q}^B_{i-1} - \pmb{q}^B_{i}) ||^2_2)
\end{equation}
where $\pmb{W}_R$ and $\pmb{W}_B$ represent weight matrices, and the cost function exclusively penalizes the traveling distance for both the mobile base and the manipulator joints between two configurations. 

\section{Simulation}\label{sec:sim}
This section presents the results from extensive simulations that evaluate the proposed \ac{smmp} framework. The simulations demonstrate the effectiveness of generating coordinated base-arm-object motion in the proposed A-Space, quantitatively compared to several baselines. Additionally, an object rearrangement task highlights the advantages of the \ac{akr}-based planning domain design in simplifying the task planning by reducing unnecessary action predicates of separately moving the mobile base and the arm. An ablation study on a complex, 18-step long-horizon \ac{smmp} task further examines the plan refinement algorithm. 

\subsection{Simulation Setup}
The simulated mobile manipulator platform comprises a Clearpath Husky mobile base and a Universal Robot UR5e robotic manipulator equipped with a Robotiq 2-finger gripper positioned at the mobile base's rotation center.
The mobile base is assumed to be omnidirectional during trajectory optimization. As the four wheels can be controlled independently, its trajectory is then processed by adjusting the orientation of the mobile base to match the direction of movement, and the shoulder joint of the manipulator is adjusted accordingly to ensure the correctness of the trajectory.

\begin{figure*}[t!]
    \centering
    \includegraphics[width=\linewidth]{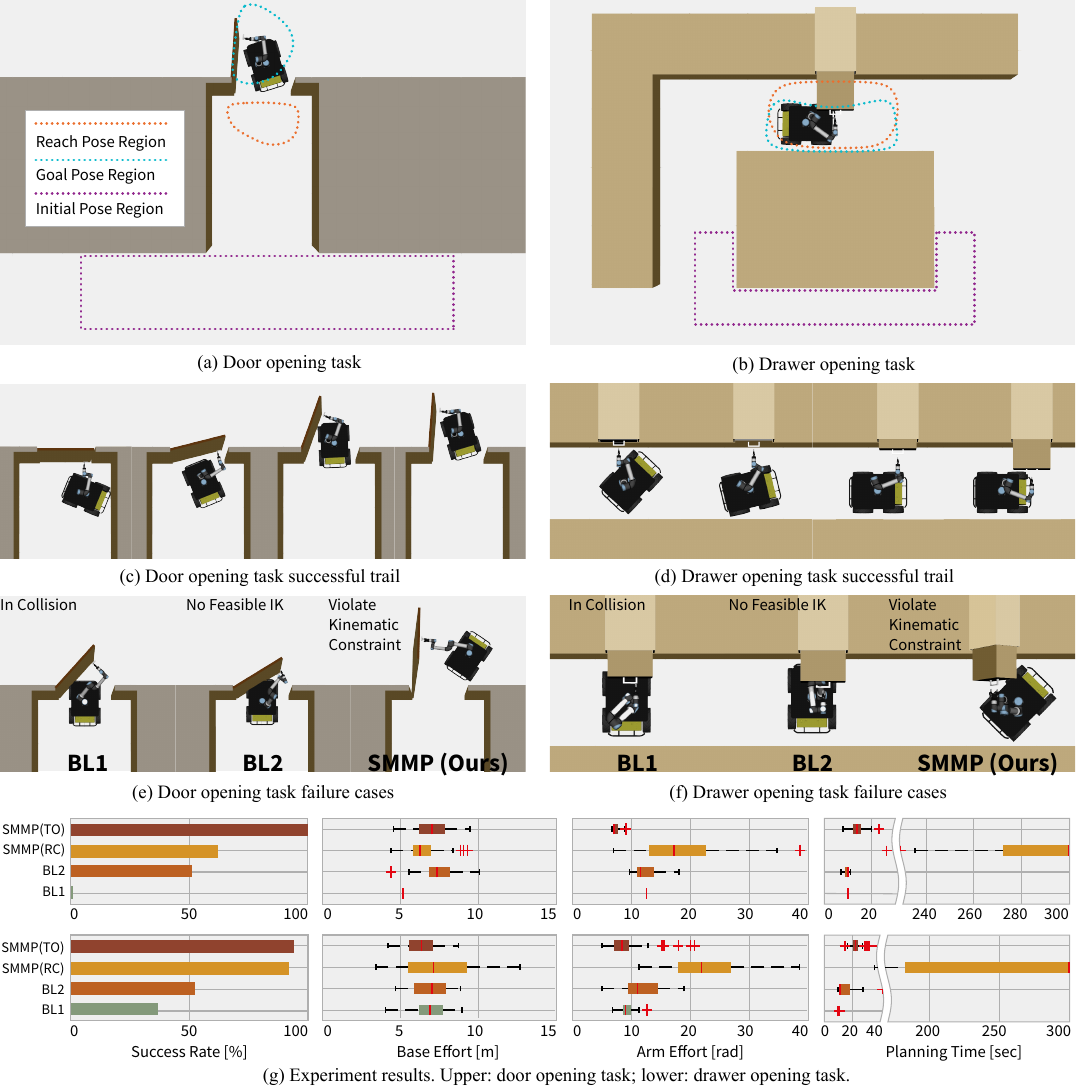}
    \caption{\textbf{Quantitative comparisons between the \ac{smmp} and three baselines in door manipulation and drawer manipulation tasks.} The robot starts from a randomized location within the purple region and (a) opens the door to a specific angle or (b) pulls out the drawer by a specific length. The orange and blue regions indicate the feasible poses that must be given to the mobile base for baselines. The regions are empirically found to guarantee a valid IK solution for the robot. (c) and (d) are successful trials of two tasks, respectively, and (e) and (f) are typical failure cases of baseline methods and the proposed \ac{smmp} method. (g) The planning success rate and the box plots with kernel density plots of the base effort, arm effort, and planning time of the four methods in these two tasks. }
    \label{fig:baseline}
    \vspace{-12pt}
\end{figure*}

\subsection{Comparisons with Baselines}\label{sec:exp:baseline}
We developed two mobile manipulation scenarios to evaluate the benefits of  \ac{smmp} as compared to approaches that treat the base and arm separately. The first task, depicted in \cref{fig:baseline}(a), is to approach a door and open it by pushing. The door has a single revolute joint and is located at the end of a corridor. The second task, illustrated in \cref{fig:baseline}(b), involves reaching a drawer in a confined kitchen space and opening it by pulling its prismatic joint. The initial position of the robot is randomly selected from within the shaded purple region.

In addition to our \ac{smmp} framework, referred to as \ac{smmp}+TO (\ie, trajectory optimization), we introduce three alternative setups to solve the above two mobile manipulation tasks for comparing the performance. To compare with a sampling-based constrained motion planner, we adopt the well-known RRT-Connect method~\cite{kuffner2000rrt} from the \ac{ompl}~\cite{sucan2012open,kingston2019exploring} to solve the constrained motion planning problems formulated by \ac{smmp}, referred to as \ac{smmp}+RC. To compare the \ac{smmp}-based approaches with typical non-\ac{smmp} approaches, we introduce two additional baselines that independently compute trajectories for the mobile base and manipulator. Baseline 1 (BL1) utilizes A$^\star$ to search for a feasible mobile base path and subsequently smooth through trajectory optimization. The arm pose is then determined by solving the inverse kinematics from the door handle to the mobile base at each way-point. Building upon BL1, Baseline 2 (BL2) further optimizes the poses of the manipulator and manipulated object at each way-point for collision avoidance. 

Notably, our \ac{smmp}-based approaches (\ac{smmp}+TO and \ac{smmp}+RC) only needs to specify one task goal: the desired door angle or the desired drawer length to open. In contrast, non-\ac{smmp} approaches (BL1 and BL2) require specification of the pose of the mobile base when reaching the doorknob as well as after having opened the door, as the base and the manipulator are planned individually. We compute the mobile base's intermediate poses by sampling from feasible regions that are empirically determined (\ie, for mobile base poses in this region, the existence of an arm pose to grasp the handle is guaranteed); see the orange areas in \cref{fig:baseline}(a)(b) for reaching, and the blue areas for final poses.

We evaluate the planning results using four criteria: (i) \textit{success rate} as the percentage of task completion without violating constraints, (ii) the \textit{base's effort} as the total base travel distance, (iii) the \textit{arm's effort} as the sum of each joint's cumulative angular displacement throughout task execution, and (iv) the \textit{planning time}. The results are summarized in \cref{fig:baseline}(g). Planning the base, arm, and manipulated object separately (BL1) results in a success rate of 1\% for opening doors and 36\% for opening drawers. The primary cause of the failures in BL1 is collisions between the mobile manipulator and the door or drawer, as shown in \cref{fig:baseline}(e)(f). Although implementing robot-object collision checks to refine motions (BL2) enhances the success rate to 51\%, the proposed \ac{smmp}-based approaches still outperform the non-\ac{smmp} approaches. Failure cases of BL2 primarily arise from kinematic constraint violations, such as the end-effector disengaging from the handle, due to the absence of feasible IK solutions caused by the bulky mobile base. This highlights the necessity of simultaneous coordination of the base, arm, and object, as illustrated in the BL2 failures shown in \cref{fig:baseline}(e)(f).

The \ac{smmp}+TO and \ac{smmp}+RC both generate feasible trajectories for the given task, with \ac{smmp}+TO achieving higher success rates than \ac{smmp}+RC. \ac{smmp}+TO produces more efficient trajectories in terms of shorter base and arm travel distances. Typically, sampling-based motion planners struggle to incorporate kinematic and safety constraints, necessitating extra effort to accommodate additional kinematic constraints~\cite{kingston2019exploring}. \cref{fig:baseline}(e)(f) also demonstrate failure cases of violating kinematic constraints. The typical failure mode of \ac{smmp}+RC is that the planner fails to find a feasible solution within the allowable time budget (300 seconds).

Our comparative experiments suggest that \ac{smmp}-based approaches are better suited for complex mobile manipulation tasks by jointly optimizes base–arm–object movements. Moreover, trajectory optimization proves more effective for solving \ac{smmp}-based motion planning problems due to the intricate constraints involved. However, this comes at the cost of increased planning time compared to non-\ac{smmp} baselines, as \ac{akr} introduces a higher \ac{dof} compared to robot kinematics.

\subsection{Analysis on Efficiency Improvement in Task Planning}\label{sec:res:task}
\begin{figure}[t!]
    \centering
    \includegraphics[width=\linewidth]{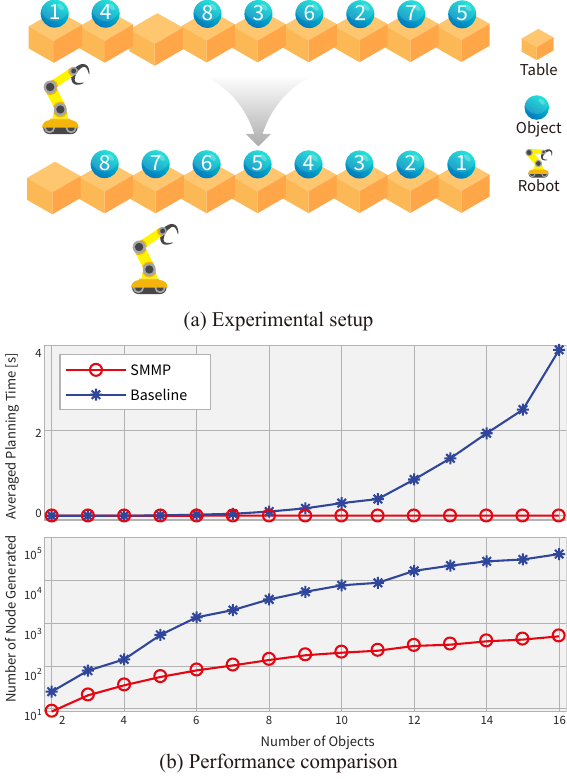}
    \caption{\textbf{\ac{akr}-based domain specification improves the task planning efficacy.} (a) An example setup of rearranging 8 objects on 9 tables; one table can only support one object. (b) The \ac{akr}-based domain specification allows a solver to search for a feasible plan for tasks involving re-arranging 2 to 16 objects in significantly less time while generating fewer nodes in search (\ie, less memory).}
    \label{fig:exp-task}
    \vspace{-16pt}
\end{figure}

By treating the robot base, arm, and object to be manipulated as a whole, the design of the task planning domain based on the A-Space perspective can offer greater efficiency. We use an object-arrangement task as an example to quantitatively evaluate the improvement offered by the A-Space perspective, where the robot rearranges $m$ objects on $m+1$ tables in a sorted order while satisfying the constraint that each table can support only one object. \cref{fig:exp-task}(a) shows a typical example of the initial and goal configuration of this task with $m=8$ objects. Our \ac{pddl} implementation, built on top of \ac{akr} (see \cref{sec:supp:additional} for details), uses fewer predicates and supports more abstract action representations compared to domain definitions that decouple base and arm movements (\eg,~\cite{garrett2020pddlstream,yang2024guiding}). Specifically, such separated base-arm domains typically require: (i) additional predicates to represent the mobile base's state, resulting in more state predicates; (ii) an extra action explicitly dedicated to base movement; and (iii) more action parameters for manipulation planning.

\begin{figure*}[t!]
    \centering
    \includegraphics[width=\linewidth,trim={0cm 0cm 0cm 0cm},clip]{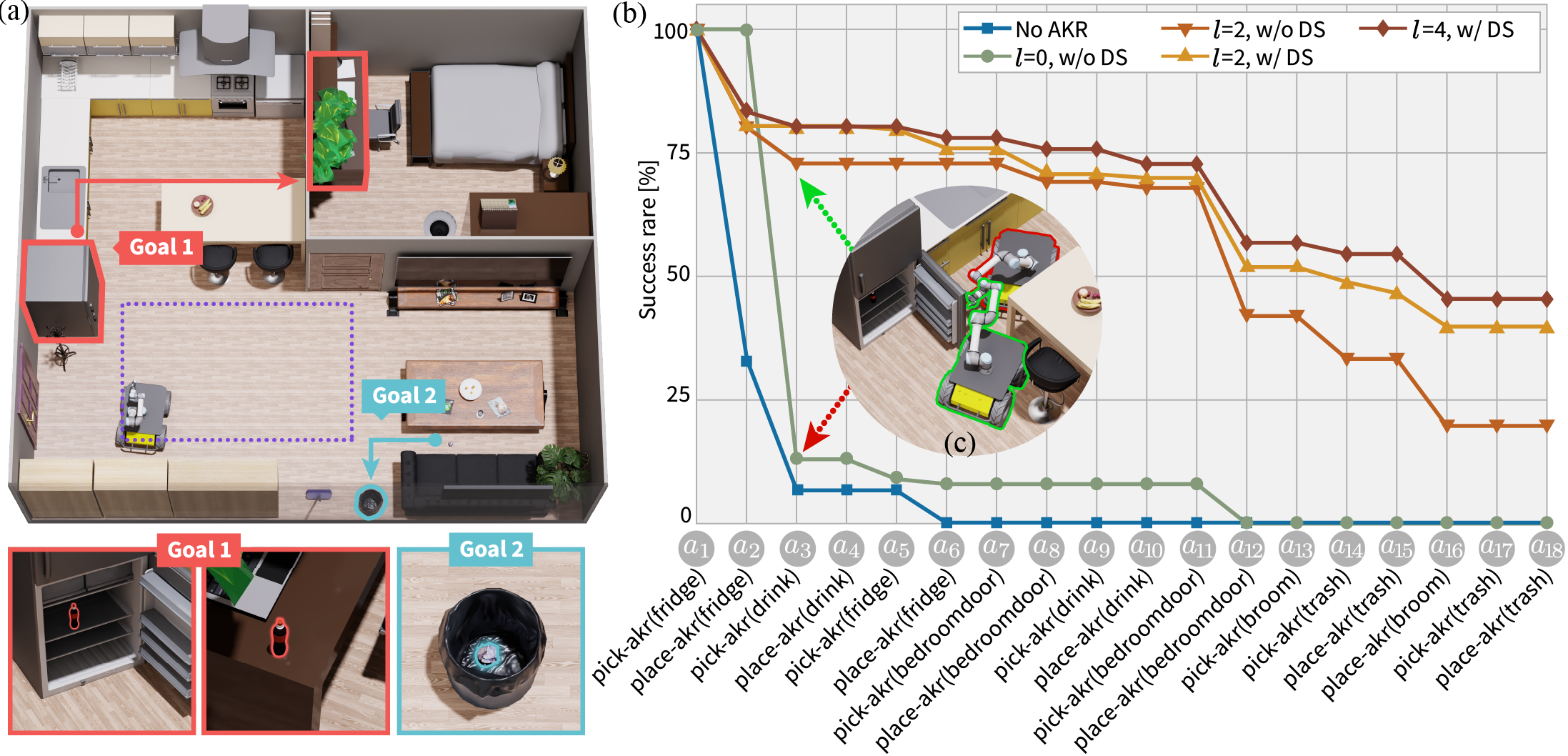}
    \caption{\textbf{Ablation studies of plan refinement in \ac{smmp} with a simulated household environment.} (a) The robot is tasked to place a drink on the desk in the bedroom and dispose of an object in the trash can. (b) The entire task consists of 18 actions, all represented by the two symbolic actions. The proposed \ac{akr} against a baseline method and different settings of plan refinement. (c) Two possible robot start configurations that influence the feasibility of subsequent actions.}
    \label{fig:results_long_sim}
    \vspace{-4pt}
\end{figure*}

To produce a task plan, we adopt the \ac{pddl} solver from~\cite{Muisepddlsolver} which employs a hybrid strategy combining a Serialized Iterative Width (SIW) search-based planner and a Best First Search-based planner, BFS(f)~\cite{lipovetzky2012width}. We use \ac{pddl} version 2.2 throughout all task planning formulations presented in this paper. In this study, we ran 50 trials for each setup; see the results summarized in \cref{fig:exp-task}(b). 
As task complexity grows, the planning time and the number of nodes generated during the search (\ie, memory usage) increase much more slowly using the \ac{akr}-based approach, as compared to separated base-arm domains, which exhibit an exponential rise. This is because non-\ac{akr}-based task planning requires more action operators to accomplish each task, resulting in greater search depths. If there are, on average, $N$ nodes generated at each search depth level, and a solution is found at depth $D$, the total nodes generated is $N^D$. The \ac{akr}-based approach requires fewer action operators, resulting in shallower search depths and substantially reducing both the number of expanded nodes and the frequency of backtracking during task planning. In a task involving rearranging 16 objects across 50 trials, we observed an average of 76\% reduction in search depth with the \ac{akr}-based task planner, leading to a significant enhancement in planning efficiency alongside a reduction in memory usage.

Taken together, the results of this study demonstrate that task planning based on \ac{akr} significantly reduces the need for wide and deep exploration in the search process. This improvement not only improves the efficiency and reduces memory usage of task planning, but also holds promise for reducing the number of motion planning calls required in broader \ac{tamp} frameworks.

\subsection{Ablation Studies of Plan Refinement in \texorpdfstring{\ac{smmp}}{} } \label{sec:res:ablation}

We conducted further ablation studies to evaluate how the plan refinement algorithm impacts the execution success rate in long-horizon tasks through simulation. In \cref{fig:results_long_sim}(a), the robot is assigned the task of bringing a drink from the fridge to the bedroom desk (Goal 1) and fetching trash located between the sofa and the coffee table before depositing it into the trash can (Goal 2). Notably, the robot has to temporarily place the drink on the dining table to free its gripper before opening the bedroom door (\circled{$a_4$}-\circled{$a_9$}), and must also use the broom to collect the trash because the space within which it is located is smaller than the robot's base (\circled{$a_{13}$}-\circled{$a_{17}$}). This poses significant challenges to the reliability of the generated 18-step task and motion plan. Please refer to \cref{sec:supp:additional} for details on the task planning domain designed for this problem.

In the study, we recorded the robot's cumulative planning success rate at each step in the sequential mobile manipulation over the 18 steps across 5 settings:

\begin{itemize}[leftmargin=*,noitemsep,nolistsep,topsep=0pt]
    \item \texttt{Non-\ac{smmp}}: The execution trajectories are generated using Baseline 2 (BL2) as described in \cref{sec:exp:baseline}, which plans the robot's base and arm separately without incorporating the proposed \ac{akr}.
    \item \texttt{$l$=0, w/o DS}: The end configuration for the robot's next action is randomly sampled, without employing plan refinement (as $l$=0) and down-sampling.
    \item \texttt{$l$=2, w/o DS}: The end configuration for the robot's next action is determined using the plan refinement algorithm \textit{without down-sampling}. A total of 3 subsequent actions (including the current one) are considered.
    \item \texttt{$l$=2, w/ DS}: The end configuration is determined using the proposed plan refinement algorithm with \textit{down-sampling enabled}. The same 3 subsequent actions (including the current one) are considered to allow direct comparison with the previous setup.
    \item \texttt{$l$=4, w/ DS}: A total of 5 subsequent actions (including the incoming action) are considered, with the other settings remaining the same as in the previous setting.
\end{itemize}

In each trial, the mobile manipulator is randomly positioned within the purple-dotted region depicted in \cref{fig:results_long_sim}(a); a total of 100 trials are conducted to obtain the cumulative success rate. To ensure a fair comparison, in addition to the initial and goal states of the environment, both the \ac{smmp} and non-\ac{smmp} methods received identical manually defined grasping poses for all movable objects, though not the corresponding robot configurations.   

\begin{figure*}[t!]
    \centering
    \includegraphics[width=\linewidth]{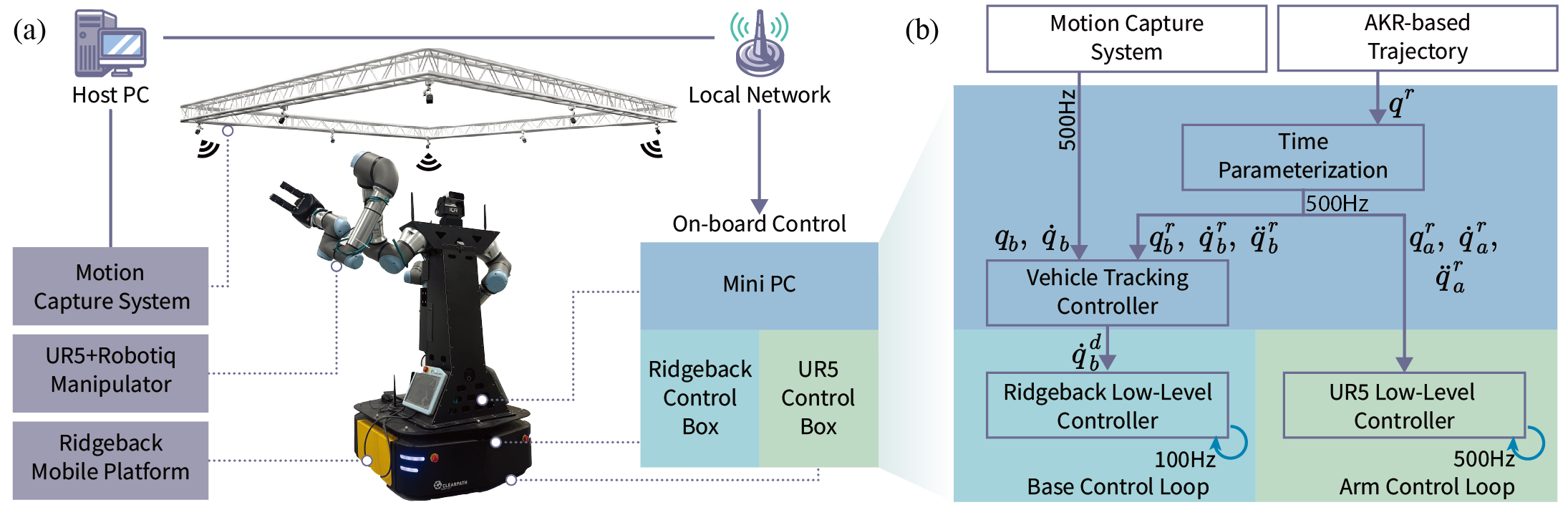}
    \caption{\textbf{The system diagram for the mobile manipulator platform.} (a) The mobile manipulator's hardware configuration and communication diagram. (b) The control diagram of the mobile manipulator platform.} 
    \label{fig:tieta_system}
\end{figure*}

In \cref{fig:results_long_sim}(b), the cumulative success rates for each set, as the task progresses, underscore the importance of considering future actions in long-horizon tasks. Without planning in the A-Space, the motion planner faces particular challenges at (\circled{$a_2$}), when opening the fridge door in a confined space. When no plan refinement algorithm is applied (\ie, $l=0$, w/o DS), the success rate significantly drops at \circled{$a_3$} because the opened fridge door and the kitchen table obstruct the robot's path to picking up the drink, as illustrated in \cref{fig:results_long_sim}(c). Without considering future actions, the robot could easily trap itself during execution in a crowded scene (\ie, select a poor end configuration as in \circled{$a_2$}). By anticipating the actions of picking up the drink and placing it in \circled{$a_3$} and \circled{$a_4$} (\ie, $l=2$, w/o DS), the robot avoids getting trapped by choosing the green end configuration instead of the red one, as in \cref{fig:results_long_sim}(c), despite this trajectory being less efficient and harder to compute at the current step, as indicated by a slight drop in motion planning success rate.

Furthermore, finding end configurations for each action using the down-sampling method improves the success rate by excluding robot configurations sharing similar geometric properties ($l=2$, w/ DS \vs $l=2$, w/o DS), so it is more likely to find a feasible goal within the computational budget (max 5 retries are allowed for each action). Looking ahead to longer horizon tasks (\ie, $l=4$, w/ DS), one could further improve the success rate, albeit at the cost of increased computational effort. 
In summary, the plan refinement algorithm significantly improves the motion planning success rate by selecting action goals at each step that take into account the feasibility of future actions, making this method better suited to long-horizon tasks.

\section{Experiments}\label{sec:exp}
This section presents the real-world robot experiments that show that the framework can generate coordinated whole-body motions for mobile manipulation across various scenarios, solve challenging long-horizon tasks, and generalize to different robot platforms and tool-use tasks analogous to the body schema theorem.

\begin{figure*}[t!]
    \centering
    \includegraphics[width=\linewidth]{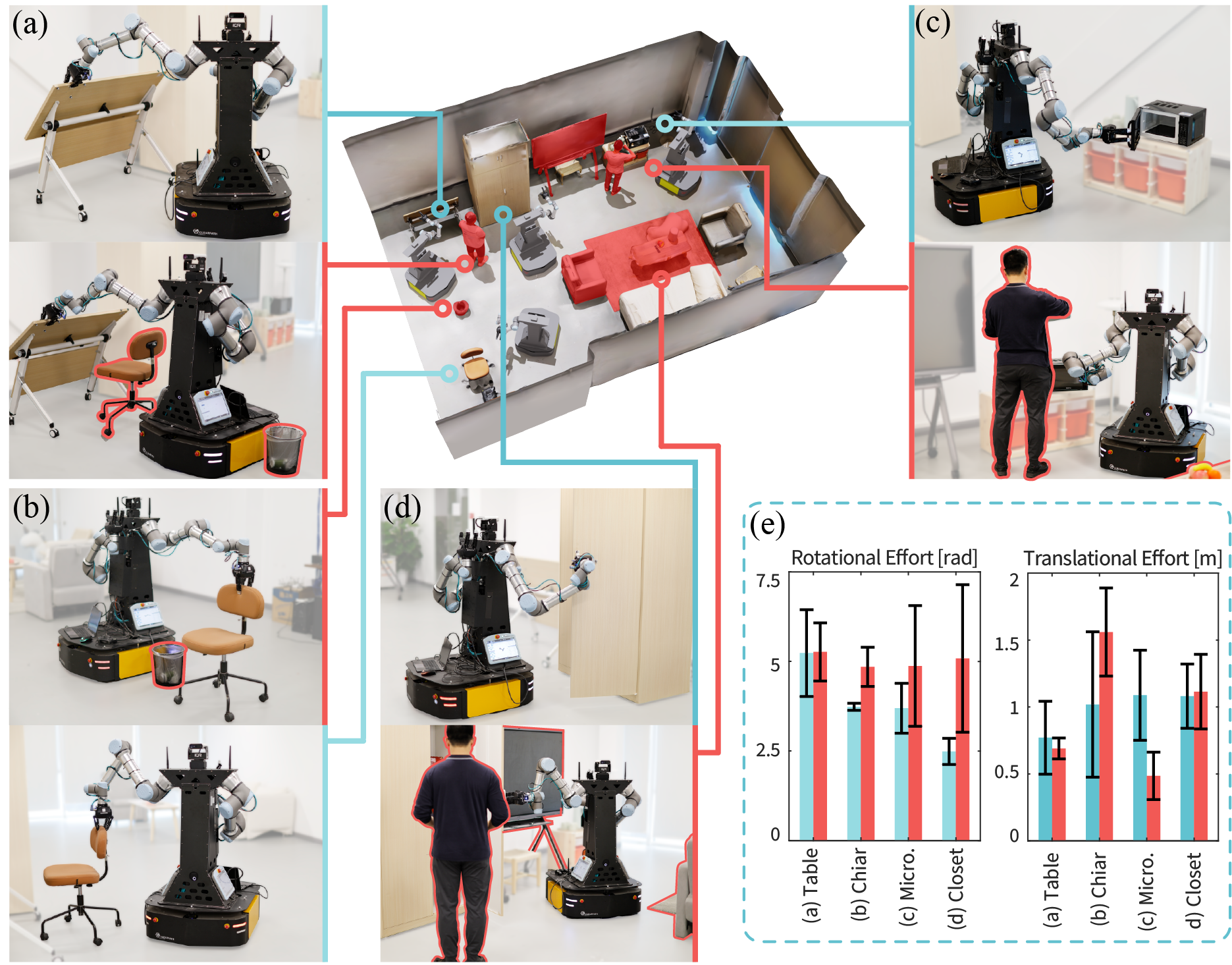}
    \caption{\textbf{Robot performance in mobile manipulation with articulated objects.} The robot generates coordinated base-arm-object motions for (a) unfolding the table, (b) placing the chair, (c) opening the microwave, and (d) opening the closet. The robot consistently maintains coordination in manipulations, even in the face of more challenging situated constraints due to the obstacles shaded in red. (e) The translational and rotational traveling distance.}
    \label{fig:results}
    \vspace{-12pt}
\end{figure*}

\subsection{Robot Platform}\label{sec:exp:platform}
In this article, we evaluated the proposed \ac{smmp} method on three robot platforms with different structures.

\textbf{The mobile manipulator platform} consists of dual Universal Robot UR5e robotic manipulators equipped with Robotiq 3-finger grippers installed on a Clearpath Ridgeback omnidirectional mobile platform. The on-board computational hardware is a mini PC with Intel Core i7-10700 CPU. Task and motion generation are performed on a host PC equipped with an AMD Ryzen9 5950X CPU. For perception, we utilize the \ac{mcs}, operating at $500~Hz$, to track object poses and the mobile base's position and orientation. \ac{mcs} data is processed on its host PC and transmitted to the onboard mini PC via a local network. It is worth noting that a single UR5e manipulator serves all tasks in our setup.

Our on-board control processes involve several steps. Firstly, the \ac{mcs} tracks the mobile base's 3D pose and objects' 6D poses in the physical environment, which are then transmitted to the host PC along with the manipulator configuration (UR5e joint values in this case) to update the \ac{akr} state. Next, the proposed \ac{smmp} framework updates the \ac{akr} structure according to the previous action and then generates the trajectory based on the current action goals and A-Space. Subsequently, the host PC sends the planned trajectory to the mini PC for time parameterization, adhering to hardware constraints. The time-parameterized reference trajectory is then sent to the corresponding base and arm controllers concurrently. For manipulator control, we utilize the built-in trajectory follower of the UR5e. For the mobile base control, we develop a custom PID-based velocity tracking controller to generate velocity control signals for trajectory tracking. \cref{fig:tieta_system} is a schematic diagram of the platform.

\textbf{The aerial manipulator platform} setup is similar to the mobile manipulator platform, the major difference being the flying vehicle control system. The high-level controller communicates with the \ac{mcs} through Ethernet for feedback and outputs the desired attitude and thrust of each thrust generator~\cite{su2021nullspace}. These commands are transmitted through Crazy Radio PA antennas (2.4 GHz) to the Crazyflie 2.1 control boards, where double-loop PID controllers are implemented for $500~Hz$ low-level control with onboard IMU feedback. \cref{sec:supp:uam_system} provides more details of the system.

\begin{figure*}[t!]
    \centering
    \includegraphics[width=\linewidth]{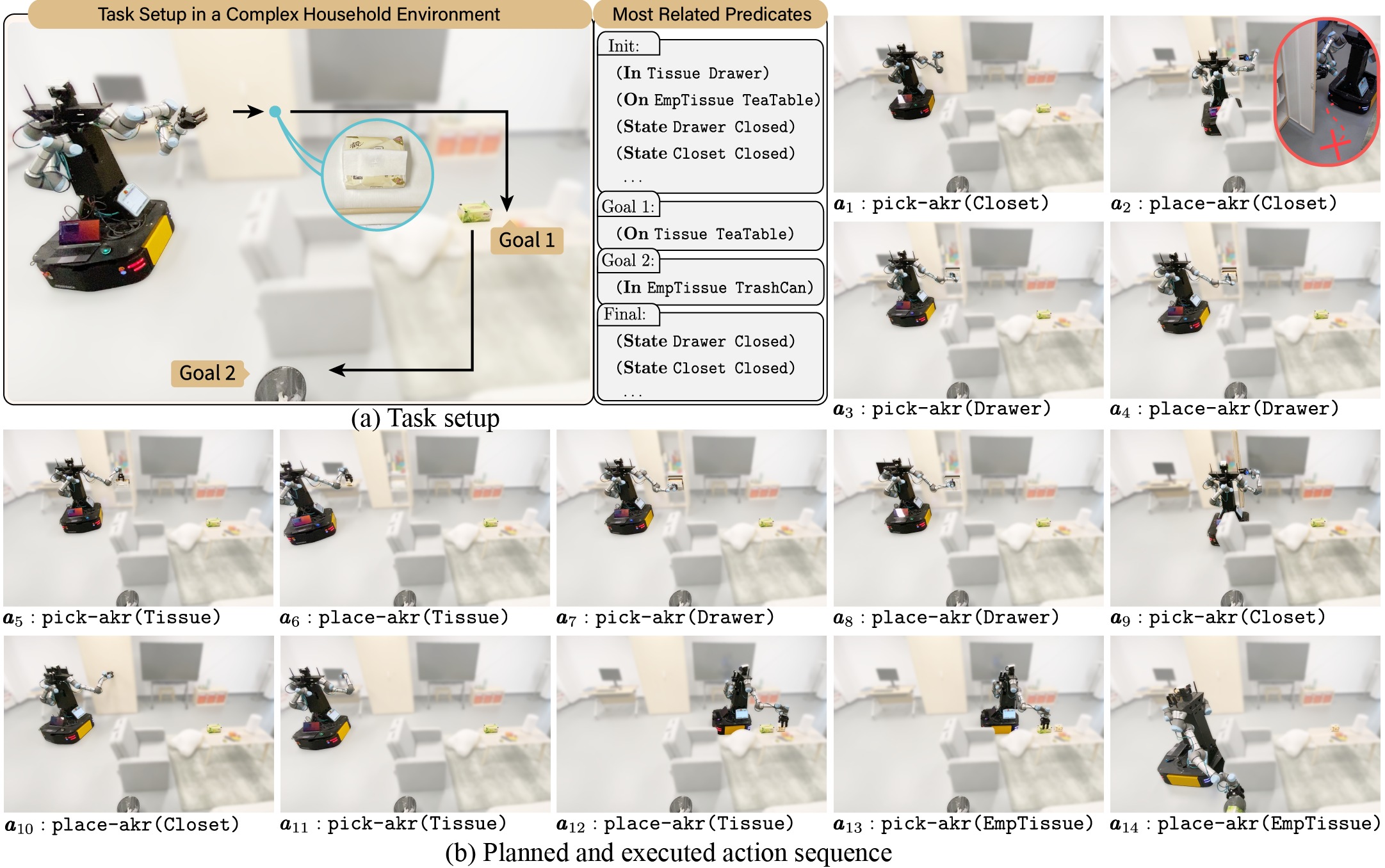}
    \caption{\textbf{Performance of the robot in a sequential mobile manipulation task requiring multiple types of action.} (a) The robot is tasked to dispose of the empty tissue box and replace it with a new one. The task goals and environment states are defined using the \ac{pddl}. (b) The robot can solve the 14-step task using only two action operators based on the \ac{akr}-based task planner. The feasible trajectories over this long-horizon task are produced accordingly, with plan refinement based on the sequence of the actions.}
    \label{fig:results_long}
    \vspace{-3pt}
\end{figure*}

\subsection{Coordinated Whole-body Trajectory Generation} \label{sec:res:coordinate}

In a real household environment featuring diverse everyday objects with distinct articulation, we showcase the robot's adept execution of various mobile manipulation tasks through coordinated whole-body motions generated by the proposed method. Snapshots in \cref{fig:results} depict the robot performing four typical household tasks: (a) unfolding a flip-top table with a horizontal revolute axis, (b) rolling a chair that can move around on the floor plane (\ie, 2D displacement), (c) opening a microwave and (d) opening a closet, both involving a vertical revolute axis, with the bulky closet door requiring more sophisticated motion coordination. By formulating motion planning problems in A-Space one naturally accommodates both robot and object movements, resulting in successful and efficient task execution. This advantage is particularly evident in \cref{fig:results}(e), in which each manipulation task is rendered increasingly complex by the presence of new obstacles (highlighted in red), and more sophisticated obstacle avoidance strategies, therefore, become necessary. Notably, for this kind of task, motion planning using in A-Space shares the same objectives and goal states, differing only in the constraint that specifies obstacle configurations in the surrounding space (see \cref{sec:planning:motion} for details). 

We further evaluated the motion planner's performance in terms of the efficiency of the trajectories it computes. By repeating the planning for each scenario in \cref{fig:results}(a)-(d) five times (with random start locations) and executing the planned trajectories on physical robots, we reported statistics on base efforts and arm joint efforts measured by trajectory lengths in \cref{fig:results}(e), to assess execution efficiency. In general, surrounding obstacles could constrain navigation, compelling the robot to compensate by increasing arm movements, and leading to notably higher joint effort when manipulating the chair, microwave, and closet.

\begin{figure*}[t!]
    \centering    
    \includegraphics[width=\linewidth]{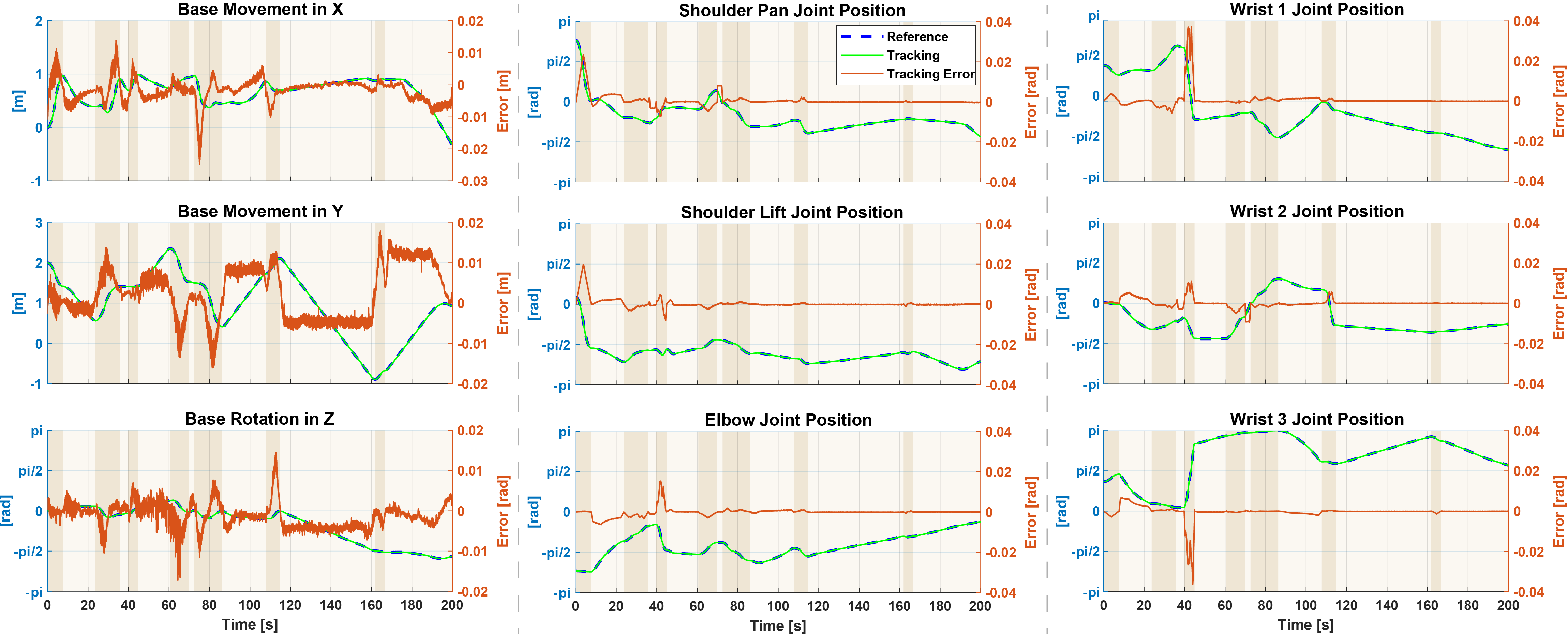}
    \caption{\textbf{Tracking performance in a real-world \ac{smmp} task.} The figure shows the reference trajectory generated from the proposed \ac{smmp} method, alongside the actual mobile manipulator's trajectory obtained from \ac{mcs} and robot joint feedback, and the tracking error. The duration of each \texttt{pick-akr} action is highlighted with a darker background.}
    \label{fig:tieta_exp_traj}
    \vspace{-5pt}
\end{figure*}

\begin{figure*}[t!]
    \centering
    \includegraphics[width=\linewidth]{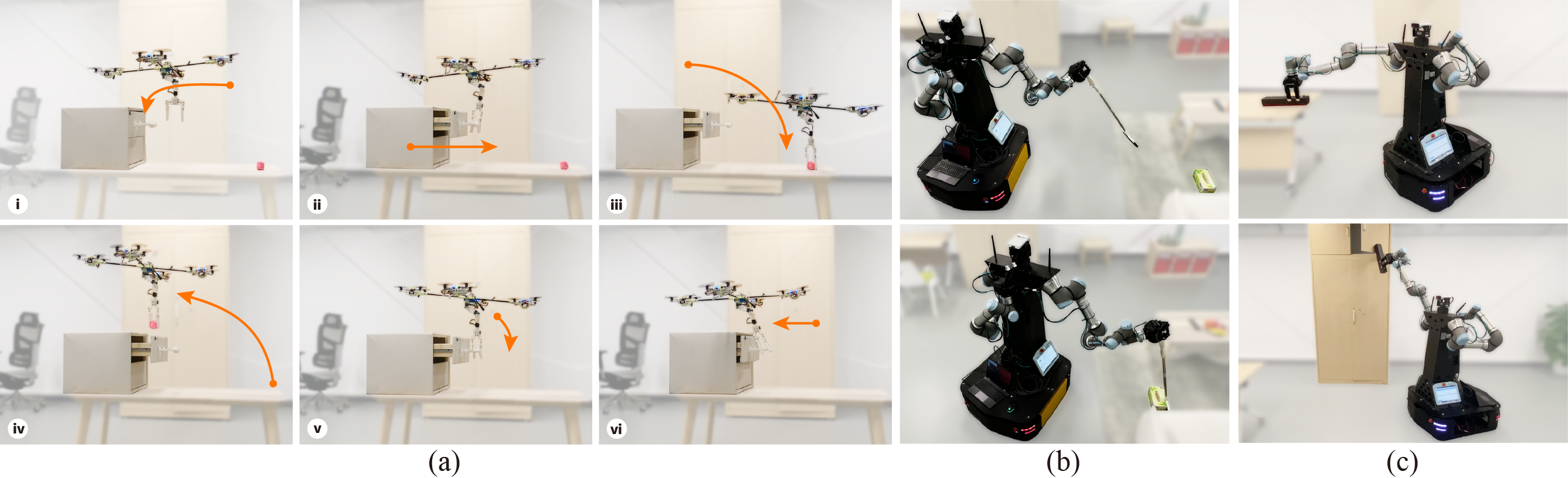}
    \caption{\textbf{Applications of the \ac{akr} for different robots and robotic tool-use tasks.} (a) Planning sequential aerial manipulation tasks for an over-actuated \ac{uam}. By abstracting the over-actuated UAV's flight as a 6-\ac{dof} floating mobile base and combining the kinematics of the 3-\ac{dof} manipulator, the proposed \ac{smmp} framework is applied to solve the task of placing an object into the drawer. (b) The \ac{akr} allows the robot to utilize manipulated objects as tools to fetch the litter with a broom and (c) to close the unreachable closet door with a stick.}
    \label{fig:results_robot}
    \vspace{-12pt}
\end{figure*}

\subsection{Sequential Mobile Manipulation Planning} \label{sec:res:smmp}

\cref{fig:results_long}(a) showcases a series of the robot in a conducting complex, long-horizon sequential mobile manipulation task. The task objectives assigned to the robot are: (Goal 1) retrieving a new tissue box from a closet drawer, placing it on the tea table, and (Goal 2) disposing of the empty tissue box in the trash can. As this task involves interactions with various structures, such as the closet with a revolute joint, the drawer with a prismatic joint, and other rigid objects while navigating through confined 3D spaces, the entire task execution consists of 14 distinct actions. Our method successfully addresses this \ac{smmp} challenge, demonstrating progress at three levels. 

At the task level, the difficulties in specifying the planning domain and the computational cost in solving the task are reduced, since only two action operators are required. These improvements are quantitatively evaluated in a simplified setup (\cref{sec:res:task}). Throughout execution, the task planner correctly determines the sequence of actions, such as opening the closet before accessing the drawer, and vice versa when closing them. This suggests that our task planning setup faithfully describes the scene and the associated state transitions.

At the motion level, motion planning problems instantiated from symbolic actions are effectively solved, resulting in well-coordinated movements of the robot's base, arm, and manipulated object during mobile manipulation (see tracking performance in \cref{fig:tieta_exp_traj}). These whole-body motions enable the robot to perform interactive tasks within confined spaces. 
 
At the goal level, selecting a robot configuration that aligns with the motion planner's goal at the end of each action is crucial for the overall success of the task. For instance, in action $\pmb{a}_2: \texttt{place-akr(closet)}$, the robot could attempt to open the closet door from either the left or right side. Our plan refinement algorithm successfully accounts for the subsequent action of pulling out the drawer ($\pmb{a}_3$) and therefore selects a robot configuration such that the door is opened from the left, thus avoiding potential obstructions from the nearby carpet and the closet door that would have just been opened. 

These three advancements collectively enable a robot to conduct \ac{smmp} tasks proficiently. \cref{fig:tieta_exp_traj} illustrates the tracking performance of the mobile manipulator. The results demonstrate that the proposed \ac{smmp} method can generate executable trajectories that are trackable by the physical robot.

\subsection{Versatility of the Proposed SMMP Framework}\label{sec:res:schema}

The advantages of formulating \ac{smmp} from the A-Space perspective extend beyond specific robots or tasks. Since kinematic relationships can characterize various patterns of a robot's movements and a wide range of task goals, our \ac{smmp} framework can be extended to other non-traditional setups in mobile manipulation. As illustrated in \cref{fig:results_robot}(a), by applying \ac{smmp} to an over-actuated \ac{uam}, which consists of an over-actuated \ac{uav} and a 3-\ac{dof} manipulator, we open up new horizons in sequential multi-step aerial manipulation. Unlike fundamentally stable ground robots, aerial robots have to prioritize their own safety - an increasingly challenging task when interacting with the surrounding environment. We, therefore, implemented a hierarchical control framework for the aerial manipulator to stabilize itself and track desired trajectories~\cite{su2021nullspace,su2023sequential}.

Planning in the A-Space also allows a robot to incorporate external objects as body extensions for non-prehensile manipulation or tool uses. \cref{fig:results_robot}(b)(c) illustrate two robotic tool-use tasks modeled and computed by \ac{smmp}. In the first task, the robot utilizes a broom to sweep away litter located between the tea table and the sofa, which the robot cannot approach directly. In the second task, the robot plans to use a stick to close the closet's upper door which is unreachable by its manipulator. These results demonstrate that our framework is not limited to a specific setting; it can be applied to different robot embodiments and has the potential to significantly expand a robot's capabilities by incorporating grasped objects as tools, a crucial step forward in open-world, task-rich environments.

\section{Discussion and Conclusion}\label{sec:conclusion}

\subsection{Key Findings}
\textbf{Coordinated Robot-scene Motion in Human Environments:} Through a series of single-step and multi-step mobile manipulation tasks in \cref{sec:res:ablation} and \cref{sec:res:smmp}, we demonstrated the effectiveness of the proposed \ac{smmp} framework in generating coordinated robot-scene motions in long-horizon tasks. This type of motion coordination in various settings is crucial for robots operating in human environments that have been primarily designed with bipedal locomotion. Indoor scenes are typically organized to meet human activities, but can be too confined and cluttered for mobile manipulators to navigate and interact with~\cite{wang2023rearrange}. Although prior work has improved the robustness and efficiency of planning algorithms in confined and cluttered environments~\cite{berenson2008optimization,han2020towards,nam2021fast}, the absence of coordinated whole-body motion still fundamentally limits robotic capabilities in many tasks (see supplementary video for examples). After integrating the manipulated object's kinematics with that of the robot, planning in the A-Space can facilitate a general and efficient formulation for robots needing to manipulate a variety of objects with whole-body motions, irrespective of the robot's own morphology. This is evidenced by our experimental results in \cref{fig:results} and \cref{fig:results_robot}, where the proposed approach successfully generates coordinated base-arm-object trajectories across a variety of scenarios involving a ground mobile manipulator and an aerial manipulator. The robots effectively handle interactions with a range of articulated furniture, like doors and drawers, and achieve significantly higher success rates compared to non-\ac{smmp} methods, as quantified in \cref{sec:exp:baseline}.

\paragraph*{Integrated Representation for Sequential Tasks}
Success in solving \ac{smmp} tasks relies heavily on the fluent execution of each single-step action. One notable advantage of framing motion planning problems based on the integrated robot-scene representation, \ac{akr}, is the clarity of directly defined goal configurations (\ie, the action parameters) at each step in terms of object states, eliminating the need to specify robot states as part of the goal. For instance, in the task of opening a microwave in \cref{fig:results}(c), the goal is to set the microwave door to a particular angle. The robot's pose and mobile base location are less critical, as they are optimized to adhere to the situational constraints (\ie, the human's location) during motion planning, which have been incorporated into the A-Space. Consequently, in \cref{fig:results}(e), when the level of confinement for the same task increases due to obstacles, the \ac{akr}-based planner readily adapts base-arm coordination and produces different trajectories to achieve the same task goal. In contrast, planning methods that treated robot base and arm movements separately require separate goal specifications and action predicates for each component; see \cref{fig:exp-task}. While this approach may be computationally efficient in motion planning, it presents challenges in coordinating movements, especially when dealing with environmental constraints imposed by external objects. As shown in \cref{fig:baseline}(g), baselines that separate base and arm planning suffer a significant drop in task success rates due to the lack of coordination between their respective end configurations. This misalignment becomes particularly problematic in complex \ac{smmp} setups, where the iterative nature of \ac{tamp} causes failed motion planning attempts to trigger frequent backtracking and replanning, ultimately leading to failures.

\subsection{Limitations and Future Directions}
\textbf{Efficient Planning for Responsive Operation:} As reported in \cref{sec:exp:baseline}, A-Space planning times, produced via trajectory optimization, are notably faster than sampling-based methods but still exceed those of baselines due to the incorporation of additional \acp{dof}. While the proposed \ac{smmp} framework can generate flexible and coordinated trajectories within confined spaces, it may be less suitable for applications that require responsive operation. A recent study by Sundaralingam~\etal introduces a potential solution by parallelizing trajectory optimization computations on GPUs~\cite{sundaralingam2023curobo}. Their approach demonstrates promising results, speeding up times by a factor of 60. By integrating this GPU-accelerated motion generation library with our \ac{smmp} framework, we achieve high-\ac{dof} dexterity and responsive operation~\cite{li2024dynamic}. Furthermore, our approach demonstrates its applicability across a series of realistic scenes adapted from iThor~\cite{kolve2017ai2} (see \cref{sec:supp:additional} for details), reinforcing its potential for real-world deployment.

\textbf{Obtaining Scene Kinematics:} We also acknowledge that the success of the proposed \ac{smmp} framework relies heavily on precise knowledge of scene kinematics, which may not always be available in unstructured environments. Recent advancements in computer vision have enabled the reconstruction and inference of part-level relations among objects with articulation~\cite{mo2019partnet,bokhovkin2021towards,zhang2023part}, offering the potential to acquire object kinematics from vision alone~\cite{han2021reconstructing,han2022scene}. Still, the precision required for manipulation exceeds the current state-of-the-art in computer vision techniques. Integrating tactile feedback at the robot's end-effectors (\eg, vision-based tactile sensors) and employing advanced adaptive controllers could enhance robot execution in scenarios with uncertain object kinematics due to perception noise. Moreover, by leveraging readily available environment datasets with known kinematics such as~\cite{kolve2017ai2}, the proposed \ac{smmp} framework can serve as an effective data generation platform, addressing the persistent challenge of high-quality data collection in learning-based manipulation research~\cite{kroemer2021review}.

\textbf{Interacting with Scenes:} Perceiving human-made scenes and the objects within them naturally guides the actions of agents~\cite{gibson1950perception,gibson1966senses}, forming the foundations for accomplishing complex tasks. However, existing approaches typically focus on capturing 2D or 3D occupancy information for obstacle avoidance during navigation or pick-and-place manipulation. To tackle longer-horizon tasks, it is crucial to incorporate \textit{actionable} information, such as the actions that entities in the scene can perform and the physical constraints they impose, into robot planning~\cite{han2022scene,jiao2022sequential}. Identifying what information can be considered actionable and beneficial for subsequent manipulation tasks is a fundamental challenge addressed in this article. Our investigation into \ac{smmp} suggests that kinematics could serve as a key bridging stone between perception-based scene understanding and control-based manipulative robot actions.

\begin{figure*}[t!]
    \centering
    \includegraphics[width=\linewidth]{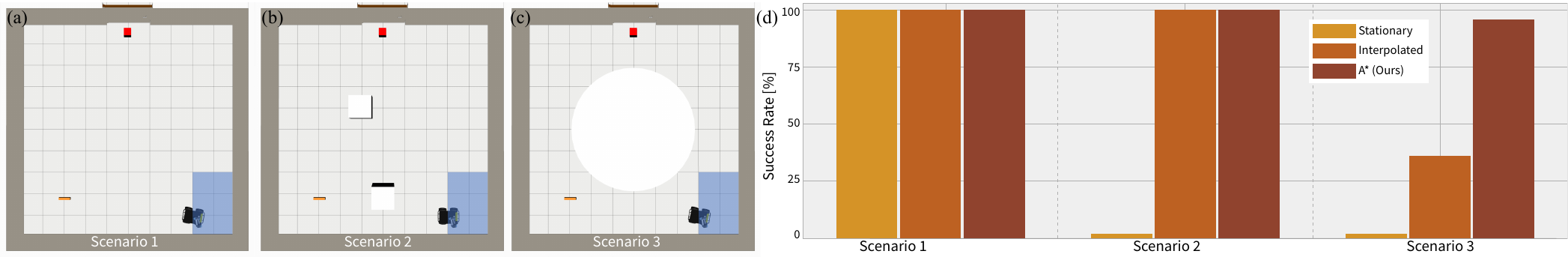}
    \caption{\textbf{Comparisons of motion planning on \ac{akr}s by different trajectory initialization methods.} (a)-(c) The experimental scenarios in increasing complexity. The robot's initial pose is uniformly sampled within the blue region; it is tasked to pick up the stick and use it to reach the red cube. (d) The proposed A$^\star$-based trajectory initialization has the highest success rates (almost always) in generating a feasible plan. In comparison, the \textit{Stationary} method fails to generate feasible plans in Scenarios 2 and 3. Similarly, the \textit{Interpolated} method struggles in Scenario 3 ($30\%$ success rate).} 
    \label{fig:trajinit}
    \vspace{-12pt}
\end{figure*}

\begin{figure*}[ht!]
    \centering
    \includegraphics[width=\linewidth]{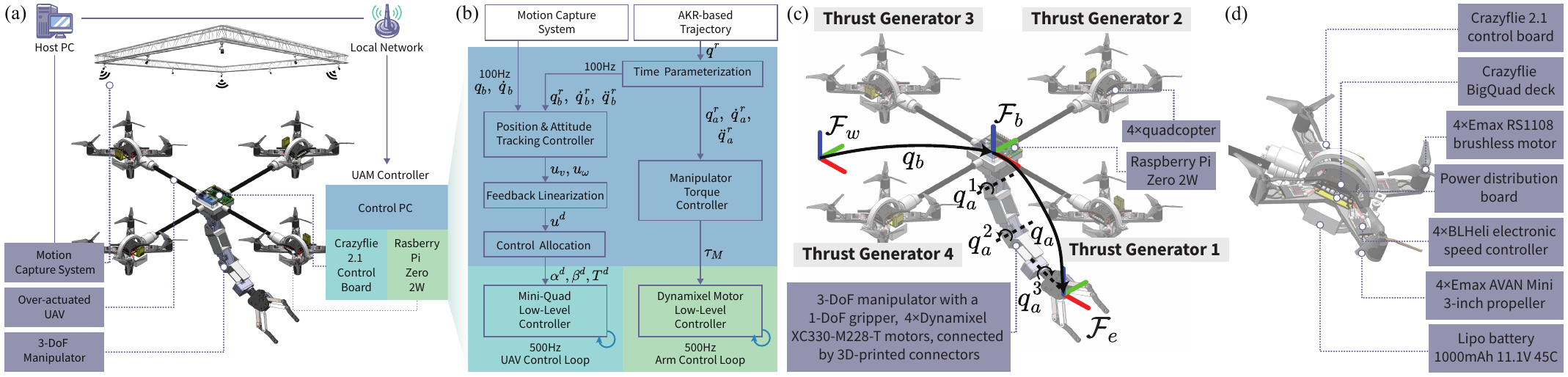}
    \caption{\textbf{The system diagram for the aerial manipulator platform.} (a) The aerial manipulator's communication diagram. (b) The control diagram of the platform. (c) The design of the aerial manipulator platform, and (d) the omnidirectional thrust generator.} 
    \label{fig:supp:uam_system}
    \vspace{-12pt}
\end{figure*}

\subsection{Conclusion}
In this article, we introduced the concept of the \acf{akr}, which integrates scene kinematics into the robot's own model to construct a unified Augmented Configuration Space (A-Space) for solving sequential mobile manipulation tasks. We developed a tri-level planning framework that combines \ac{pddl}-based task planning, trajectory optimization, and plan refinement, and validated it extensively through both simulation and real-world experiments. Our results demonstrate the framework's effectiveness in generating coordinated whole-body motions, even in confined spaces with articulated objects, and its ability to execute complex tasks involving up to 14 sequential actions without interruption. As kinematics offers a general representation of constrained motion beyond robotic morphology alone, the proposed \ac{akr} and A-Space framework holds strong promise for broad application across diverse robot platforms and challenging manipulation scenarios.

\appendix

\subsection{Trajectory Initialization}\label{sec:supp:trajinit}

We implement two trajectory initialization baselines~\cite{magyar2019timed}:

\begin{enumerate}[leftmargin=*,noitemsep,nolistsep,topsep=0pt]
    \item \textbf{Stationary:} The trajectory $\pmb{q}_{1:T}$ is initialized by way-points $\pmb{q}_t$ that are the same as the initial pose $\pmb{q}_{1}$.
    \item \textbf{Interpolated:} The trajectory $\pmb{q}_{1:T}$ is initialized by way-points that are linearly interpolated between the initial pose $\pmb{q}_1$ and the goal pose $\pmb{q}_T$.
\end{enumerate}

Next, we investigate how different trajectory initialization methods affect the planning results in three scenarios; see \cref{fig:trajinit}(a)-(c). The robot's task is to pick up the rigid stick and use it to reach a target indicated by the red cube. This task consists of two steps: i) navigate to the stick and pick it up, ii) navigate to and reach the target with the stick. The three scenarios designed for evaluation are increasing in complexity: no obstacle (\cref{fig:trajinit}(a)), two small obstacles (\cref{fig:trajinit}(b)), or a much larger one (\cref{fig:trajinit}(c)). Experimental results reported below are the average of 50 different initial poses, each with 10 times.

A successfully optimized trajectory is a converged result without violating any constraints (\eg, collisions). \cref{fig:trajinit}(d) compares success rates. When the environment is clean (Scenario 1), even the simplest \emph{Stationary} trajectory initialization method performs well. When there is additional complexity introduced by the obstacles (Scenario 2), the \emph{Stationary} method deteriorates, whereas the \emph{Interpolated} method still maintains a high success rate. When the navigable space is significantly reduced (Scenario 3), only the proposed \emph{A$^\star$}-based initialization method can consistently perform well to generate feasible plans. Taken together, experimental results indicate that combining the proposed \emph{A$^\star$}-based initialization with the optimization-based motion planner can well handle the challenging motions that require combining navigation and manipulation in cluttered space with obstacle avoidance.

\subsection{Unmanned Aerial Manipulator Platform}\label{sec:supp:uam_system}
The Unmanned Aerial Manipulator (UAM) platform consists of an over-actuated omnidirectional flying vehicle and a 3-\ac{dof} robotic manipulator~\cite{su2023sequential}. The flying vehicle integrates four omnidirectional thrust generators, each built with a generic quadcopter (Crazyflie 2.1 control board) and a 2-\ac{dof} passive gimbal mechanism~\cite{yu2021over}, enabling independent position and attitude tracking capability. The robotic manipulator comprises three serial rotational \acp{dof} and a parallel gripper. Four Dynamixel XC330-M228-T motors actuate the manipulator, while a Raspberry Pi Zero and a Dynamixel U2D2 converter are fitted to the flying vehicles to receive wireless control commands. \cref{fig:supp:uam_system} is a schematic diagram of the platform.

\subsection{Additional Materials}\label{sec:supp:additional}

The project page\footnote{\hyperlink{https://aug-kin-rep.github.io}{https://aug-kin-rep.github.io}} for this article hosts supplementary materials that could not be included in the main manuscript due to space constraints. These materials include:

\textbf{\ac{pddl} files:} Domain files for the simulation tasks described in \cref{sec:res:task,sec:res:ablation,sec:res:smmp},  along with the corresponding problem files specifying the initial and goal states, are provided. These are released together with the off-the-shelf task planner to support reproducibility and further research.

\textbf{Extended simulation results:} We tested our \ac{akr}-based mobile manipulation planner in realistic, cluttered household scenes adapted from iThor~\cite{kolve2017ai2}. Each scene includes articulated objects selected from the PartNet-Mobility dataset~\cite{xiang2020sapien}, replacing existing objects of the same category to preserve contextual realism. Grasp poses were generated using AO-Grasp~\cite{morlans2024grasp}. These results are available on the project page. To handle the scale and complexity of this evaluation, we ported our \ac{akr}-based motion planning framework to the Curobo platform and developed a fully automated toolchain for environment setup.

\textbf{Codebase:} The codebase used in our experiments, including that for the extended simulations, is released via the project page.

\balance
\bibliographystyle{ieeetr}
\bibliography{reference}

\end{document}

%% file: config.tex
\usepackage{graphicx} \graphicspath{ {figures/} }
\usepackage{amsmath,amssymb,mathabx}
\usepackage{algorithmic}
\usepackage[ruled,vlined]{algorithm2e}
\usepackage{acronym}
\usepackage{enumitem}
\usepackage{booktabs}
\usepackage{hyperref}
\usepackage{balance}
\usepackage{xspace,setspace}
\usepackage[skip=3pt,font=small]{subcaption}
\usepackage[skip=3pt,font=small]{caption}
\usepackage[dvipsnames]{xcolor}
\usepackage[capitalise]{cleveref}
\usepackage{tabularx,colortbl,multirow,array,makecell}
\usepackage{overpic}
\usepackage{cite}
\usepackage{tikz}
\usepackage{dblfloatfix}

\newcommand*\circled[1]{\tikz[baseline=(char.base)]{
            \node[shape=circle,draw,inner sep=0.6pt] (char) {#1};}}

\makeatletter
\DeclareRobustCommand\onedot{\futurelet\@let@token\@onedot}
\def\@onedot{\ifx\@let@token.\else.\null\fi\xspace}
\def\eg{\emph{e.g}\onedot} 
\def\ie{\emph{i.e}\onedot} 
 
 \def\vs{\emph{vs}\onedot}
\def\wrt{w.r.t\onedot} 
\def\etal{\emph{et al}\onedot} \def\vs{\emph{v.s}\onedot}
\makeatother

\crefname{algocf}{alg.}{algs.}
\Crefname{algocf}{Algorithm}{Algorithms}

\crefname{algorithm}{Alg.}{Algs.}
\Crefname{algocf}{Algorithm}{Algorithms}
\crefname{section}{Sec.}{Secs.}
\Crefname{section}{Section}{Sections}
\crefname{table}{Tab.}{Tabs.}
\Crefname{table}{Table}{Tables}
\crefname{figure}{Fig.}{Fig.}
\crefname{equation}{Eq.}{Eqs.}
\Crefname{figure}{Figure}{Figure}

\makeatletter
\def\BState{\State\hskip-\ALG@thistlm}
\makeatother

\makeatletter
\renewcommand{\paragraph}{%
  \@startsection{paragraph}{4}%
  {\z@}{0ex \@plus 0ex \@minus 0ex}{-1em}%
  {\hskip\parindent\normalfont\normalsize\bfseries}%
}
\makeatother

\crefname{algocf}{alg.}{algs.}
\Crefname{algocf}{Algorithm}{Algorithms}

\definecolor{gblue}{HTML}{4285F4}
\definecolor{gred}{HTML}{DB4437}

\acrodef{dof}[DoF]{Degree of Freedom}
\acrodef{vkc}[VKC]{Virtual Kinematic Chain}
\acrodef{tamp}[TAMP]{Task and Motion Planning}
\acrodef{pddl}[PDDL]{Planning Domain Definition Language}
\acrodef{rrt}[RRT]{Rapidly-exploring Random Tree}
\acrodef{ompl}[OMPL]{Open Motion Planning Library}
\acrodef{iws}[IWS]{Iterated Width Search}
\acrodef{bfs}[BFS]{Breadth First Search}
\acrodef{ai}[AI]{Artificial Intelligence}
\acrodef{mmmp}[MMMP]{Multi-Modal Motion Planning}
\acrodef{mcts}[MCTS]{Monte Carlo Tree Search}
\acrodef{ros}[ROS]{Robot Operating System}
\acrodef{urdf}[URDF]{Universal Robot Description Format}
\acrodef{kcs}[KCS]{Key Configuration Set}
\acrodef{drl}[DRL]{Deep Reinforcement Learning}
\acrodef{rl}[RL]{Reinforcement Learning}
\acrodef{ompl}[OMPL]{Open Motion Planning Library}
\acrodef{mcs}[MCS]{Motion Capture System}
\acrodef{akr}[AKR]{Augmented Kinematic Representation}
\acrodef{llm}[LLM]{Large Language Model}
\acrodef{lhmm}[LHMM]{Long-Horizon Mobile Manipulation}
\acrodef{smmp}[SMMP]{Sequential Mobile Manipulation Planning}
\acrodef{uam}[UAM]{Unmanned Aerial Manipulator}
\acrodef{uav}[UAV]{Unmanned Aerial Vehicle}
\acrodef{prm}[PRM]{Probabilistic Roadmap}
\acrodef{vlm}[VLM]{Vision-Language Model}